\documentclass[10pt,twocolumn,letterpaper]{article}

\usepackage{iccv}              

\definecolor{iccvblue}{rgb}{0.21,0.49,0.74}
\usepackage[pagebackref,breaklinks,colorlinks,allcolors=iccvblue]{hyperref}
\usepackage[accsupp]{axessibility}
\usepackage{bm}
\def\paperID{7585} 
\def\confName{ICCV}
\def\confYear{2025}

\title{Decoupled Multi-Predictor Optimization for Inference-Efficient Model Tuning}

\author{\fontsize{12pt}{13pt}\selectfont Liwei Luo\textsuperscript{1$\dagger$}, 
Shuaitengyuan Li\textsuperscript{1$\dagger$}, 
Dongwei Ren\textsuperscript{1}, 
Qilong Wang\textsuperscript{1\thanks{Corresponding author. $\dagger$Equal contribution.  \newline \hspace*{1.5em} E-mail: \{luoliwei, listy, rendw, qlwang\}@tju.edu.cn. \newline
\hspace*{1.5em} Project page: \href{https://github.com/TJU-sjyj/DMPO}{https://github.com/TJU-sjyj/DMPO}.
}},
Pengfei Zhu\textsuperscript{1,2}, 
Qinghua Hu\textsuperscript{1}
\\
\fontsize{9pt}{11pt}\selectfont 
\textsuperscript{1}Tianjin University, China\quad\textsuperscript{2}Low-Altitude Intelligence Lab, Xiong'an National Innovation Center Technology Co., Ltd\\
}

\begin{document}
\maketitle
\begin{abstract}

Recently, remarkable progress has been made in large-scale pre-trained model tuning, and inference efficiency is becoming more crucial for practical deployment. Early exiting in conjunction with multi-stage predictors, when cooperated with a parameter-efficient fine-tuning strategy, offers a straightforward way to achieve an inference-efficient model. However, a key challenge remains unresolved: How can early stages provide low-level fundamental features to deep stages while simultaneously supplying high-level discriminative features to early-stage predictors? To address this problem, we propose a \textbf{D}ecoupled \textbf{M}ulti-\textbf{P}redictor \textbf{O}ptimization (DMPO) method to effectively decouple the low-level representative ability and high-level discriminative ability in early stages. First, in terms of architecture, we introduce a lightweight bypass module into multi-stage predictors for functional decomposition of shallow features from early stages, while a high-order statistics-based predictor is developed for early stages to effectively enhance their discriminative ability. To reasonably train our multi-predictor architecture, a decoupled optimization is proposed to allocate two-phase loss weights for multi-stage predictors during model tuning, where the initial training phase enables the model to prioritize the acquisition of discriminative ability of deep stages via emphasizing representative ability of early stages, and the latter training phase drives discriminative ability towards earlier stages as much as possible. As such, our DMPO can effectively decouple representative and discriminative abilities in early stages in terms of architecture design and model optimization. Experiments across various datasets and pre-trained backbones demonstrate that DMPO clearly outperforms its counterparts when reducing computational cost. Particularly, DMPO with 30\% FLOPs is comparable with or even suppresses counterparts with 70\% FLOPs.

\end{abstract}    
\vspace{-1.5em}
\section{Introduction}
\label{sec:intro}
\vspace{-2mm}

\begin{figure}
    \centering
    \includegraphics[width=0.4\textwidth]{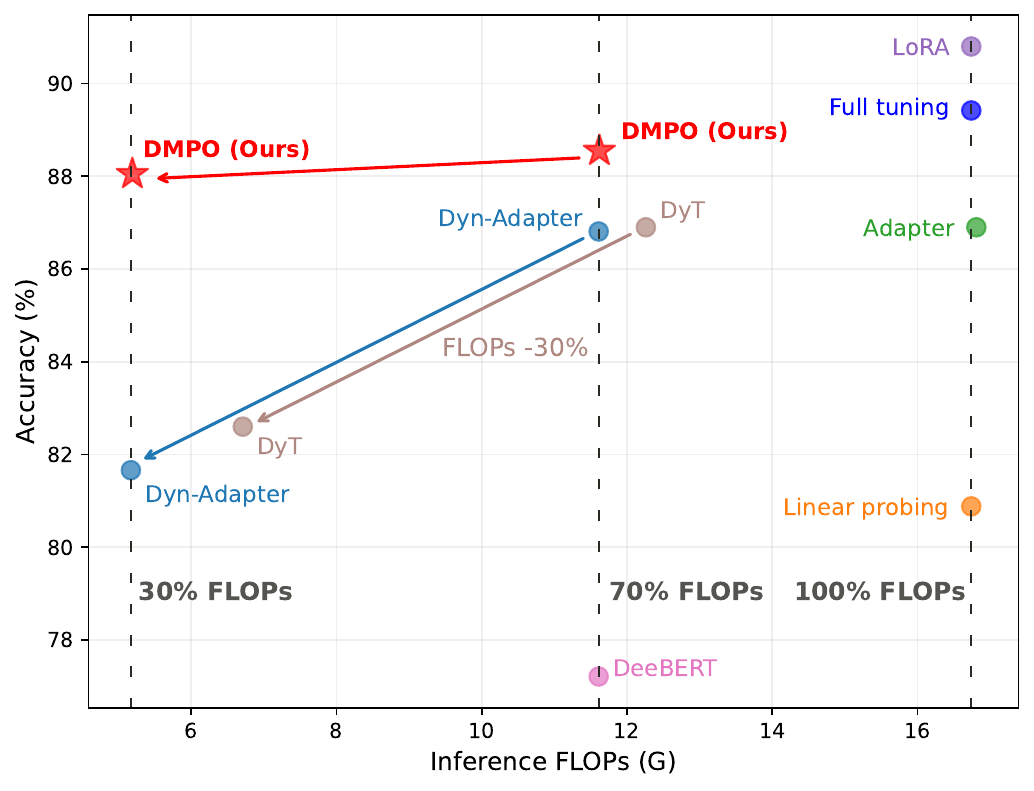}
    \setlength{\abovecaptionskip}{2pt} 
    \setlength{\belowcaptionskip}{2pt}
    \caption{Comparison of various methods in terms of inference FLOPs and average accuracy on CIFAR-100 and five FGVC datasets, with corresponding results presented in \cref{tab:fgvc}.}
    \label{fig:intro}
\end{figure}

Recent years have witnessed remarkable advancements of large-scale deep neural networks such as Vision Transformers (ViT) \cite{dosovitskiy2020vit,liu2021swin} pre-trained on large-scale datasets in various vision tasks. To fully exploit their potential, fine-tuning large-scale pre-trained models is a common practice for downstream applications. Although parameter-efficient fine-tuning (PEFT) methods such as Adapter~\cite{houlsby2019adapter}, LoRA~\cite{hu2022lora}, and VPT~\cite{jia2022vpt} are tuning-efficient, they may introduce additional inference costs due to the inclusion of fine-tuning modules. Meanwhile, inference efficiency becomes increasingly crucial for practical deployment. 
Existing inference-efficient strategies mainly include token pruning \cite{zhao2024dyntuning,sarkarECV24BSR,liu2024sparse,devoto2024adaptivelayer} by reducing representation size, and early exiting \cite{zhang2024dynadapter,liu2020fastbert,xin2021berxit,zhang2022pceebert} by outputting prediction results with incomplete inference. However, as shown in \cref{fig:intro}, token pruning method, DyT~\cite{zhao2024dyntuning}, suffers from significant accuracy drops with decreasing FLOPs. Furthermore, negative impacts have been observed on out-of-distribution performance~\cite{lin2023spurious} when removing spurious tokens. Recently, early exiting cooperating with PEFT provides a straightforward way to achieve an inference-efficient model tuning~\cite{zhang2024dynadapter}. 

\begin{figure*}
    \centering
    \includegraphics[width=0.93\textwidth]{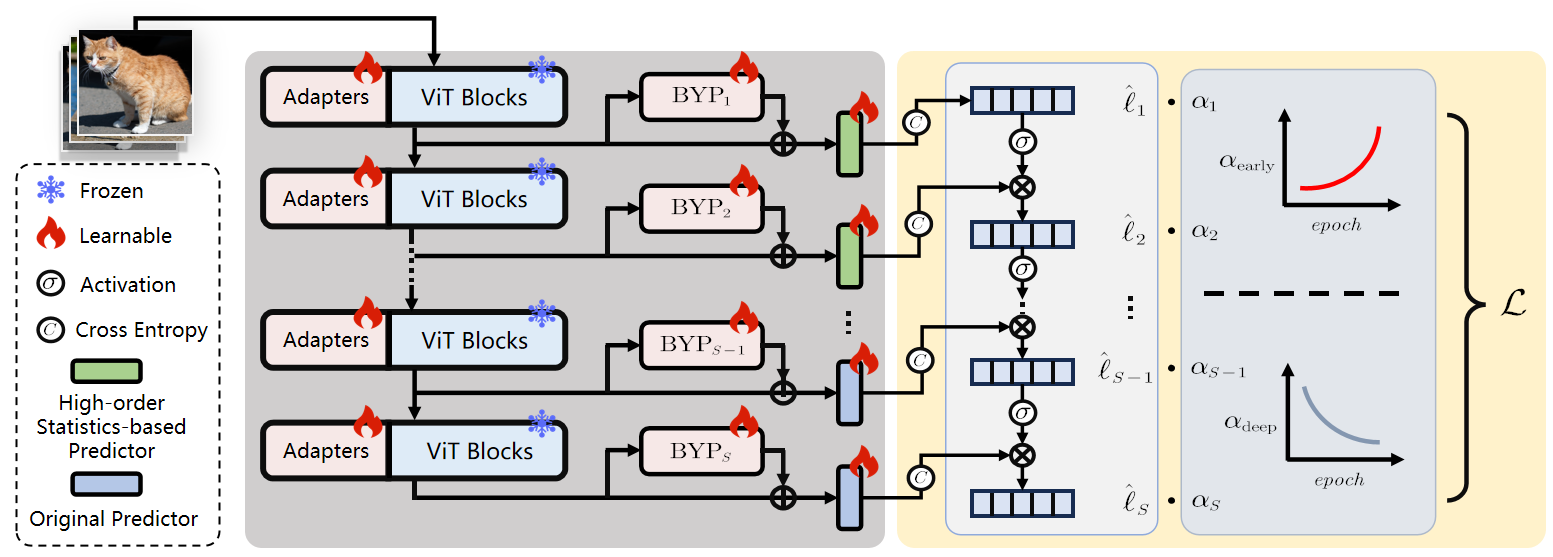}
    \setlength{\abovecaptionskip}{3pt} 
    \setlength{\belowcaptionskip}{3pt}
    \caption{Overview of our proposed DMPO method. Specifically, our DMPO consists of two components: (i) The left part of this figure is the multi-predictor architecture, inserting bypass modules, denoted as BYP, into the multi-predictors and replacing the early predictors with high-order statistics-based predictors; (ii) The right part is a decoupled optimization algorithm that adjusts multiple loss weights $\alpha$ in a two-phase manner to respectively emphasize representative ability (in the initial training phase) and discriminative ability (in the latter training phase) at early stages.}
    \label{fig:overview}
\end{figure*}
For enabling early exiting, it is crucial to enhance discriminative ability in early stages, which guarantees acceptable prediction accuracy with incomplete inference. However, it is well known that early stages primarily serve to provide fundamental features to deep stages~\cite{huang2018msd,han2023dynperceiver,lee2015deeply}, and there will exist conflict between low-level representative ability and high-level discriminative ability for early stages, as also observed in~\cite{zhang2024dynadapter}. Although advanced effort has been made in Dyn-Adapter~\cite{zhang2024dynadapter}, it still fails to address the challenge. Specifically, the architecture of Dyn-Adapter does not fully decouple role of shallow features from early stages by using identical features for subsequent stages and current predictors, while its optimization strategy overlooks discriminative features in early stages, leading to suboptimal performance under low FLOPs (refer to \cref{fig:intro}).

In this paper, we propose a \textbf{D}ecoupled \textbf{M}ulti-\textbf{P}redictor \textbf{O}ptimization (DMPO) method to decouple the low-level representative ability and high-level discriminative ability in early stages from two perspectives, as shown in \cref{fig:overview}.
First, in terms of architecture, we introduce lightweight bypass modules into the multi-predictors to encapsulate shallow features from early stages as discriminative while preserving their representative ability during transmission to deep stages. Moreover, we replace the original predictors in early stages with high-order statistics-based predictors that have stronger information extraction capabilities to enhance the discriminative power of early stages.
To reasonably train our multi-predictor architecture, we propose a decoupled optimization algorithm to allocate two-phase loss weights during tuning. In the initial training phase, we assign smaller loss weights to early stages, enabling the model to initially prioritize the acquisition of the discriminative ability of deep stages via emphasizing the representative ability of early stages. Subsequently, we progressively increase the loss weights of early stages to drive discriminative ability towards earlier stages as much as possible. Consequently, our DMPO effectively addresses the challenge of early stages, simultaneously providing low-level fundamental features to deep stages and high-level discriminative features to the early exiting predictors. As shown in \cref{fig:intro}, our DMPO achieves impressive performance and maintains comparable accuracy with significantly reduced computational cost.

To evaluate the effectiveness of our DMPO, extensive experiments are conducted on various classification benchmarks (\eg, VTAB-1K~\cite{zhai2019vtab}, FGVC datasets \cite{wah2011cub,van2015nabirds,nilsback2008flowers,gebru2017cars,khosla2011dogs,bossard2014food,maji2013aircraft}, and four variants of ImageNet \cite{recht2019imagenetv2,hendrycks2021imageneta,hendrycks2021imagenetr,wang2019imagenetsketch}) across few-shot learning and domain generalization tasks. The contributions of this work are summarized as follows: 
\begin{itemize} 
    \item 
    Through experimental analysis, we thoroughly examine the challenge of decoupling low-level representative ability and high-level discriminative ability in early stages of early exiting methods, which is crucial for achieving satisfying accuracy when requiring very low inference cost.

    \item 
    We propose DMPO from two perspectives, \ie, network architecture and decoupled optimization, to effectively decouple representative ability and discriminative ability in early stages, with a focus on driving the discriminative ability toward earlier stages as much as possible.

    \item 
    Our DMPO is applied to several downstream tasks for attaining inference-efficient models, where DMPO with only 30\% FLOPs is comparable to or even suppresses counterparts with 70\% FLOPs. 

\end{itemize}
\vspace{-2mm}
\section{Related Work}
\label{sec:formatting}
\vspace{-1mm}
\begin{figure}
    \centering
    \includegraphics[width=0.35\textwidth]{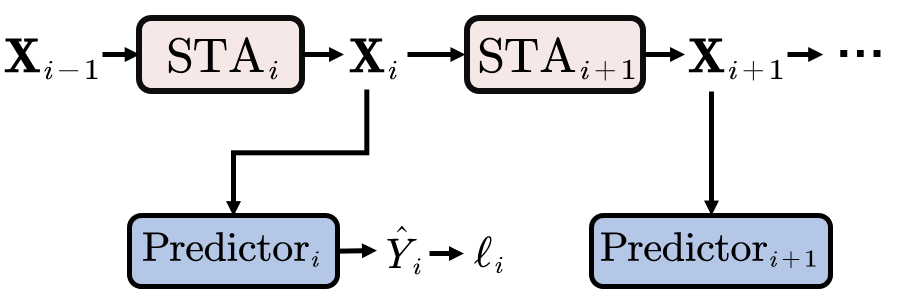}
    \caption{Schematic diagram of the early exiting network.}
    \label{fig:tra_multi}
\end{figure}
\paragraph{Efficient Fine-tuning.}
Efficient fine-tuning methods are widely applied to improve model performance with lower resources. PEFT methods \cite{houlsby2019adapter,hu2022lora,luo2023repadapter,jia2022vpt,Lian_2022_SSF, chen2022adaptformer, li2021prefixtuning, zhang2022noah, zaken2021bitfit} aim to achieve performance on downstream tasks comparable to full fine-tuning by freezing the backbone and training additional learnable parameters. However,  \cite{houlsby2019adapter,jia2022vpt,chen2022adaptformer,li2021prefixtuning,zhang2022noah} increase inference time due to the added modules. Although \cite{hu2022lora,luo2023repadapter,Lian_2022_SSF} merge these modules into the original model through re-parameterization, they still maintain the original inference speed. To enhance inference efficiency, inference-efficient fine-tuning methods have been proposed to reduce computational costs by dynamically pruning tokens during inference. DyT~\cite{zhao2024dyntuning}, for instance, discards tokens based on learned activation probabilities, while others \cite{sarkarECV24BSR,liu2024sparse,devoto2024adaptivelayer} prune tokens with weak associations to the CLS token, as determined by attention scores within the MHA module. Additionally, early exiting methods aim to reduce computational costs by selectively terminating inference for certain samples. However, most early exiting fine-tuning methods are applied to language models such as BERT \cite{zhang2022pceebert,zhou2020pabeebert,liu2020fastbert,xin2021berxit}, with limited in vision tasks~\cite{zhang2024dynadapter}.

\vspace{-1.5em}
\paragraph{Early Exiting.}
Based on the observation that simple samples can achieve accurate predictions after only early stages of a network, early exiting introduces multiple exits that allow these samples to be classified at earlier stages. Since early exiting significantly reduces computational cost during inference, it has been widely applied in the field of computer vision \cite{bolukbasi2017adaptive,yang2020ranet,huang2018msd,han2023dynperceiver,demir2024eecnn}. However, a previous study~\cite{huang2018msd} has pointed out that early stages interfere with deep stages by forcing intermediate low-level fundamental features to encapsulate high-level discriminative features, which damages the original low-level features and negatively impacts performance. To address this issue, Dyn-Perceiver~\cite{han2023dynperceiver} decouples feature extraction from early predictors by constructing feature and classification branches to process features independently. Dyn-Adapter~\cite{zhang2024dynadapter} was the first to combine early exiting with PEFT methods while leveraging the frozen pre-trained model to preserve feature extraction quality. In this work, our DMPO effectively decouples low-level representative ability and high-level discriminative ability to enhance discriminative ability in early stages, ensuring low inference FLOPs while maintaining strong performance. 
\vspace{-1em}
\section{Proposed Method}
\vspace{-2mm}
In this section, we first introduce the overview of DMPO, then describe the modifications in multi-stage predictors, and finally present the decoupled optimization algorithm.

\vspace{-0.5em}
\subsection{Overview of DMPO}
\vspace{-0.5em}

In existing early exiting methods, a pre-trained model, \eg, ViT, is divided into $S$ stages, with each stage containing $L$ blocks and a predictor. 
Through the PEFT methods, fine-tuning is performed while freezing the pre-trained backbone. \cref{fig:tra_multi} illustrates that the output feature $\mathbf{X}_{i}$ from the $i$-th stage, serves as the input to both blocks in $(i+1)$-th stage and $\textup{Predictor}_{i}$ in $i$-th stage. The output from $\textup{Predictor}_{i}$ is passed through the \texttt{Softmax} activation function to generate the classification prediction $\hat{Y}_i$, and the cross-entropy loss $\ell_i$ is computed with the corresponding target label $Y$. The loss function of existing early exiting methods can be formulated as
\begin{equation}
\setlength{\abovedisplayskip}{3pt}
\setlength{\belowdisplayskip}{3pt}
\mathcal{L} = \sum\limits_{i=1}^S \alpha_i\cdot{\ell}_{i}(\hat{Y}_i,Y),
  \label{eq:loss_early}
\end{equation}
where $\alpha_i$ is the loss weight for $i$-th stage. 

\vspace{-0.5em}
\subsubsection{Decoupling Challenge in Early Stages}
 Optimization of early exiting in Eq.~(\ref{eq:loss_early}) generally suffers from the conflict between low-level representative ability and high-level discriminative ability in early stages. To illustrate this decoupling challenge, we take Dyn-Adapter~\cite{zhang2024dynadapter} with $S=2$ as an example. Specifically, a ViT model with 12 blocks serves as the backbone, and two variants are trained on CIFAR-100~\cite{krizhevsky2009cifar}: (i) The original Dyn-Adapter; (ii) Dyn-Adapter with feature regularization, referred to as Dyn-R. Particularly, for Dyn-R, we use a mean square error loss to close the gap of feature similarity between Dyn-R and Original ViT in the 6-th block (\ie, Stage 1), aiming to enlarge low-level representative ability of early stages lying in Dyn-Adapter. More details on experiment setting are given in the supplementary material.

\begin{figure}
    \begin{subfigure}[t]{0.235\textwidth}
        \centering
        \includegraphics[width=\textwidth]{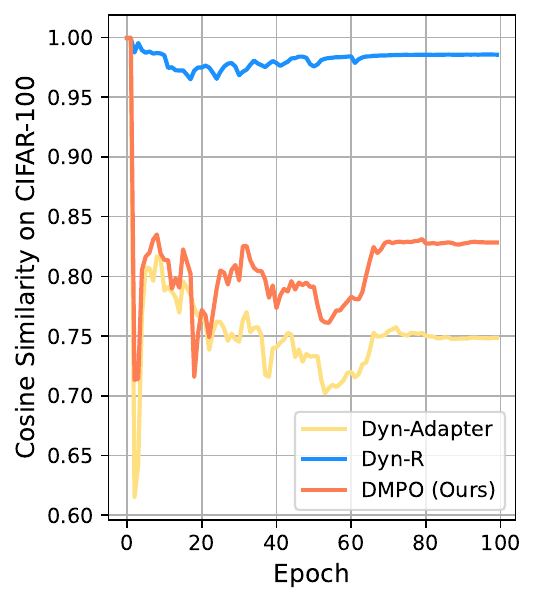}
        \caption{\footnotesize Similarity at varying epoch}
        \label{fig:similarity}
    \end{subfigure} 
    \hfill
    \begin{subfigure}[t]{0.235\textwidth}
        \centering
        \includegraphics[width=\textwidth]{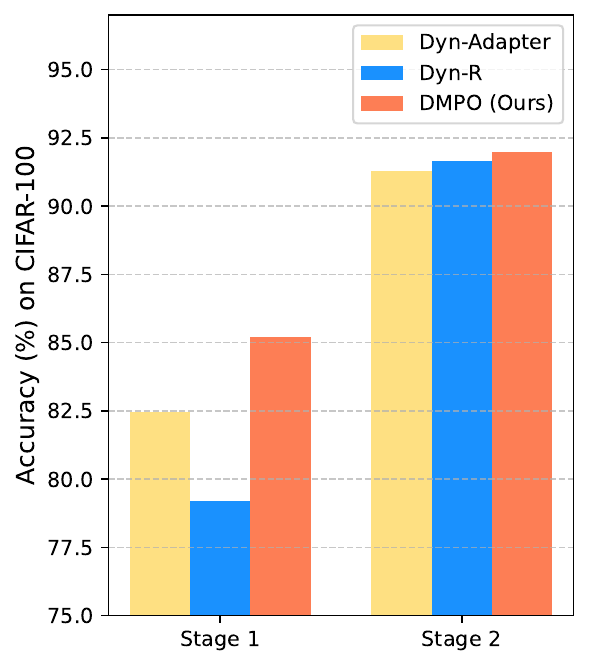}
        \caption{\footnotesize Accuracy of multi-predictors}
        \label{fig:similarity_res}
    \end{subfigure}
    \caption{Comparison of two Dyn-Adapter~\cite{zhang2024dynadapter} variants and our DMPO in terms of cosine similarity with Original ViT (Stage 1) and classification accuracy (Stage 1 and Stage 2).}
    \label{fig:simulation}
\end{figure}

The cosine similarities of features in the 6-th block (\ie, Stage 1 in Dyn-Adapter and Dyn-R) between Original ViT and two variants of Dyn-Adapter are presented in \cref{fig:similarity}, while classification accuracies at Stage 1 and Stage 2 are given in \cref{fig:similarity_res}. We observe that features of Stage 1 from Dyn-R exhibit significantly higher similarity to those of Original ViT, which primarily provides fundamental characteristics for deep stages and so leads to improved accuracy at Stage 2. However, there is a notable accuracy decrease at Stage 1 since its discriminative ability is suppressed. Although Dyn-Adapter offers a means to enhance discriminative ability at Stage 1 (\textit{w.r.t} higher accuracy), suboptimal accuracy at Stage 2 is obtained. These results clearly show the decoupling challenge lying in early stages, which remains fully unsolved in previous work~\cite{zhang2024dynadapter}. Please refer to the supplementary material for more discussion.

\subsubsection{Decoupled Multi-Predictor Optimization}

To address the challenge, we propose a new multi-predictor architecture as well as a decoupled optimization algorithm.
Once trained, the model's inference can be terminated by satisfying either a given FLOPs constraint or a desired classification confidence \cite{han2023dynperceiver,zhang2024dynadapter}.

\textbf{\textit{Multi-Predictor Architecture}}:
To resolve the challenge arising from conflict abilities, we first introduce lightweight bypass modules to encapsulate shallow features from early stages as discriminative while preserving their low-level representative ability during transmission to deep stages.
Moreover, a high-order statistics-based predictor is developed for early stages to extract richer information lying in low-level features, enhancing their discriminative ability.

\textbf{\textit{Decoupled Optimization}}:
To train our multi-predictor architecture, the total loss of our DMPO is formulated as
\vspace{-0.5em}
\begin{equation}
\setlength{\abovedisplayskip}{5pt}
\setlength{\belowdisplayskip}{5pt}
\mathcal{L} = \sum\limits_{i=1}^S \alpha_i\cdot\hat{\ell}_{i}(\hat{Y}_i,Y),
  \label{eq:loss_total}
\end{equation} 
where the loss weight $\alpha_i$ is adjusted in a two-phase manner along with training epochs. 
Moreover, to prevent adverse effects on deep stages when discriminative ability has been well acquired in early stages, we dynamically control the influence of features from early stages to deep stages, resulting in an updated loss function denoted as $\hat{\ell_i}$.

Our DMPO effectively decouples low-level representative ability and high-level discriminative ability in early stages, yielding improved performance in both early and deep stages, referring to \cref{fig:similarity_res}.

\vspace{-2mm}
\subsection{Multi-Predictor Architecture}\label{sec:mpa}
\vspace{-1mm}
\subsubsection{Bypass Module}\label{sec:bypass}
To balance early stages' ability to learn both low-level fundamental and discriminative features, we introduce lightweight bypass modules into the multi-predictors, 
\begin{equation}
\setlength{\abovedisplayskip}{3pt}
\setlength{\belowdisplayskip}{3pt}
  \widehat{\mathbf{X}}_{i} = \textup{BYP}_{i}(\mathbf{X}_{i}) + \mathbf{X}_{i},
  \label{eq:bypass}
\end{equation}
where $\textup{BYP}_{i}$ denotes the bypass module used for feature decoupling in $i$-th stage and is set by an efficient module in~\cite{hu2022lora}. These bypass modules encapsulate the original features  $\mathbf{X}_{i}$ of $i$-th stage to form new features $\widehat{\mathbf{X}}_{i}$, specifically designed for discriminative classification. As such, inherent representative ability of $\mathbf{X}_{i}$ is preserved and beneficial to discriminative ability in subsequent deep stages.

\vspace{-2mm}
\subsubsection{High-order Statistics-based Predictor}
Although $\widehat{\mathbf{X}}_{i}$ decouples abilities of early-stage features by bypass modules, simple low-level features hardly offer sufficient discriminative ability. As observed in~\cite{hp_pami}, high-order statistics-based predictor has the potential to leverage shallow network architecture to achieve promising performance. By regarding early stages in the early exiting network as a shallow network, we propose to employ high-order statistics-based predictors with confidence recalibration for enhancing discriminative ability of early stages, \ie,
\begin{equation}
\setlength{\abovedisplayskip}{6pt}
\setlength{\belowdisplayskip}{6pt}
  \hat{Y}_i = \texttt{Softmax}(\beta_i\cdot\textup{HP}_{i}(\widehat{\mathbf{X}}_{i})),
  \label{eq:higer-order}
\end{equation}
where $\textup{HP}_{i}$ is the high-order statistics-based predictor, which aims to fully leverage rich statistics of features $\widehat{\mathbf{X}}_{i}$ and provides stronger discriminative ability. To this end, $\textup{HP}_{i}$ first computes multi-head convolutional cross-covariance representations~\cite{mp} to efficiently and effectively capture second-order moments of features $\widehat{\mathbf{X}}_{i}$, which are fed into a linear classifier for discriminative prediction. Particularly, our high-order statistics-based predictor progressively decreases representation sizes in the latter stages to balance discriminative ability among different stages. Furthermore, we introduce a learnable factor $\beta_i$ for $i$-th stage to enhance classification confidence of high-order statistics-based predictor when predictions are correct. 

\begin{figure}[t]
  \centering
  \setlength{\abovecaptionskip}{1pt} 
   \setlength{\belowcaptionskip}{1pt}
  \includegraphics[width=1.0\linewidth]{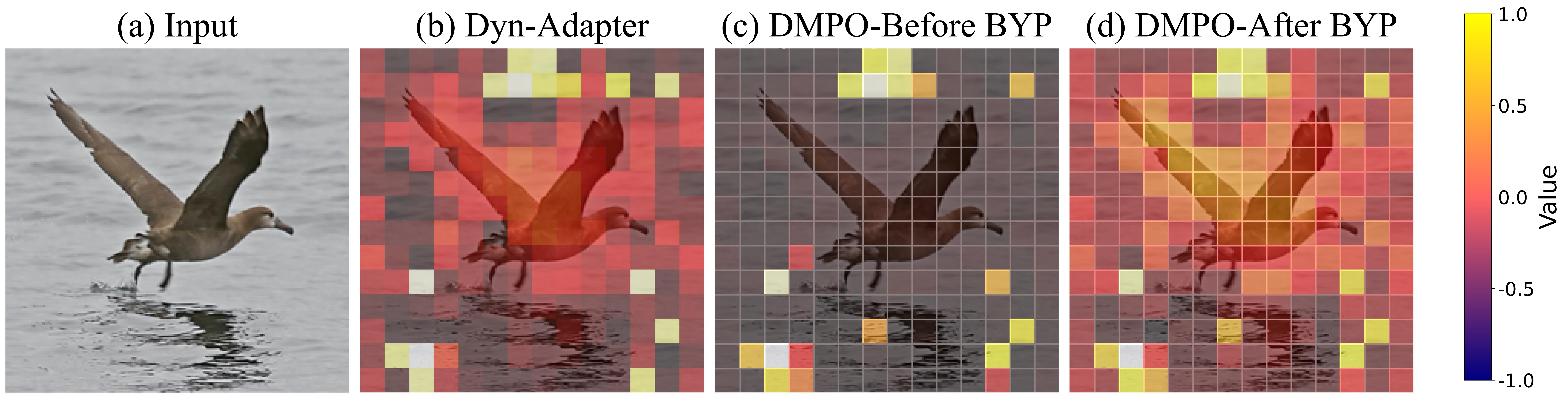}
   \caption{Heat maps of different methods at Stage 1. Dyn-Adapter propagates the feature in (b) to the next stage while feeding it into the predictor simultaneously. In contrast, DMPO propagates the fundamental features in (c) to the next stage, while transmitting the discriminative features in (d) to the predictor.
   }
   \label{fig:heatmap}
\end{figure}

\noindent\textbf{Discussion} To show how our method work for feature decoupling, we visualize heat maps of feature responses at Stage 1 in \cref{fig:heatmap}. Specifically, the features in Dyn-Adapter are confused, since they are used as low-level fundamental cues and high-level discriminative features simultaneously. On the contrary, for our DMPO, features before BYP in \cref{fig:heatmap}(c) retain their low-level representativeness and are propagated to subsequent stages, while features after BYP in \cref{fig:heatmap}(d) guided by our HP become more discriminative, thereby improving prediction accuracy at earlier stages. It may account for the effectiveness of our DMPO architecture in decoupling features.

\subsection{Decoupled Optimization}
To effectively train our multi-predictor architecture, we propose a decoupled optimization method. As shown in \cref{eq:loss_total}, it involves two key components, \ie, two-phase loss weight $\alpha_i$ allocation strategy for decoupled training and adaptive factors $\hat{\ell_i}$ for reducing inter-stage influence. 

\begin{table}[t!]
\centering
\small
\begin{tabular}{lc}
\toprule
\textbf{Method} & \textbf{Loss Weight Allocation Strategy} \\
\midrule
DeeBERT~\cite{xin2020deebert} & $\alpha_i = 1$ \\
MuE~\cite{MuE}   & $\alpha_i = {1}/{S}$ \\
CALM~\cite{calm}   & $\alpha_i = i/\sum_{j=1}^S{j}$ \\
Dyn-Adapter~\cite{zhang2024dynadapter}  & $\alpha_{\text{early}} < \alpha_{\text{deep}}$ \\
\hline
\multirow{2}{*}{DMPO (Ours)} & Initial phase: $\alpha_{\text{early}} < \alpha_{\text{deep}}$ \\
 & Latter phase: $\alpha_{\text{early}} > \alpha_{\text{deep}}$ \\
\bottomrule
\end{tabular}
\setlength{\abovecaptionskip}{2pt} 
\setlength{\belowcaptionskip}{2pt}
\caption{Comparison of different loss weight allocation strategies. Note that $\alpha_{\text{early}}$ and $\alpha_{\text{deep}}$ denote the loss weights of early and deep stages, respectively.}
\label{tab:loss_weight_allocation}
\end{table}

\subsubsection{Decoupled Training}\label{sec:alpha}
\begin{figure}
    \begin{subfigure}[t]{0.235\textwidth}
        \centering
        \includegraphics[width=\textwidth]{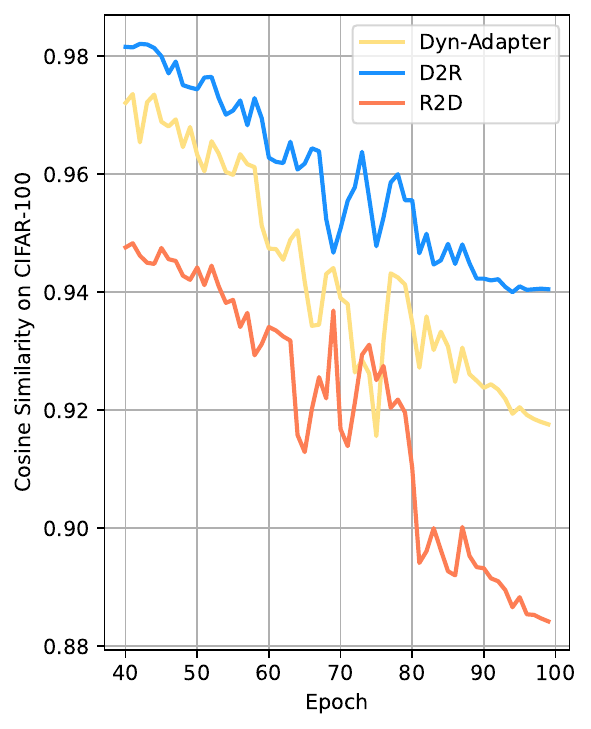}
        \caption{\footnotesize Similarity at varying epoch}
        \label{fig:feature-sim}
    \end{subfigure} 
    \hfill
    \begin{subfigure}[t]{0.235\textwidth}
        \centering
        \includegraphics[width=\textwidth]{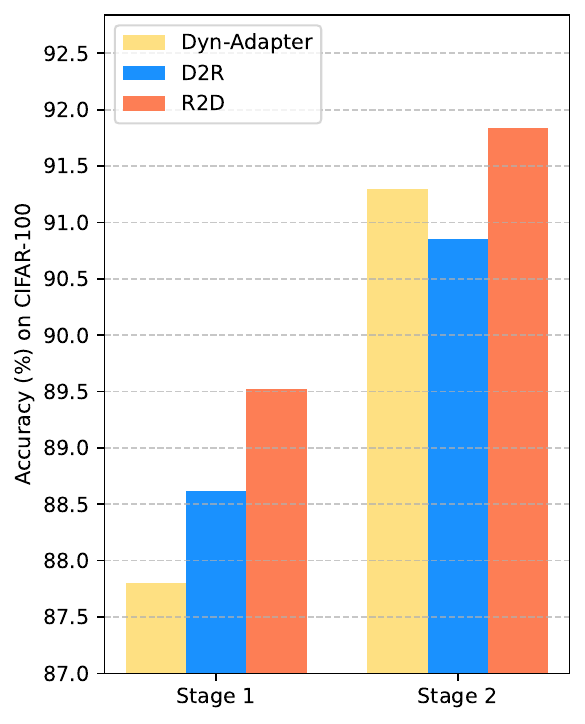}
        \caption{\footnotesize Accuracy of multi-predictors}
        \label{fig:feature-sim_res}
    \end{subfigure}
    \caption{Comparison of cosine similarity (Stage 1) and classification accuracy (Stage 1 and Stage 2) across different loss weight allocation strategies.}
    \label{fig:simulation}
\end{figure}

Although multi-stage predictors in \cref{sec:mpa} can decouple representative and discriminative abilities in terms of architecture, training by standard optimization strategy~\cite{xin2020deebert,MuE,calm,zhang2024dynadapter} (\ie,  simultaneous learning of two abilities) still leads to feature confusion and inferior performance (see \cref{fig:feature-sim_res}). Therefore, we propose a decoupled training strategy to learn representative and discriminative abilities separately, which involves two cases: (i) Generation of low-level features with early stages as primary ability is learned first, and subsequently discriminative ability is increased (namely R2D); (ii) Discriminative ability is learned first, followed by learning representative ability (namely D2R). Particularly, these training cases can be easily achieved by adjusting $\alpha$ in \cref{eq:loss_total}, as compared in \cref{tab:loss_weight_allocation}. 

Specifically, different training cases correspond to various loss weight allocation strategies: (i) For standard optimization, $\alpha_{\text{early}} < \alpha_{\text{deep}}$ is fixed throughout training (\ie, Dyn-Adapter); (ii) For R2D, the initial training phase follows $\alpha_{\text{early}} < \alpha_{\text{deep}}$ and the latter phase follows $\alpha_{\text{early}} > \alpha_{\text{deep}}$; (iii) For D2R, the initial training phase follows $\alpha_{\text{early}} > \alpha_{\text{deep}}$ and the latter phase follows $\alpha_{\text{early}} < \alpha_{\text{deep}}$.

To investigate the effect of different strategies for ability decoupling in early stages, we compare with different loss weight allocation strategies in terms of cosine similarity of features before and after BYP at Stage 1. As shown in \cref{fig:feature-sim}, R2D demonstrates a notably lower cosine similarity than others, progressively achieving better feature decoupling as training progresses.
Specifically,  initial training epochs for early stages, $\alpha_{\text{early}} < \alpha_{\text{deep}}$ is employed to guarantee representative ability, while $\alpha_{\text{early}} > \alpha_{\text{deep}}$ is utilized to enhance the discriminative ability during the later training epochs.  Therefore, we use R2D as the training strategy of our DMPO, and such a two-phase setting of $\alpha$ is robust for various tasks. For a theoretical analysis, please see Section S2. The same configuration (see \cref{sec:implementation} for details) is adopted throughout all experiments in this paper.

\vspace{-2mm}
\subsubsection{Reducing Inter-stage Influence}\label{sec:reduce_inf}
As discussed in \cref{sec:alpha}, decoupled optimization may impact the discriminative ability of deep stages as it is progressively driven towards earlier stages. To mitigate this effect, we propose an adaptive control factor $\hat{\ell_i}$ to adjust updates of features in deep stages. Specifically, when early stages can provide adequate features for discriminative classification, the control parameters will tend to freeze features in deep stages. In contrast, if early stages provide fundamental features that cannot account for precise classification, the control parameters tend to update features in deep stages.
\begin{table*}[t!]
    \centering
    \small
    \setlength{\tabcolsep}{1.5pt}
    \begin{tabular}{l@{\hspace{0.2em}}|ccccccc|cccc|cccccccc|ccc}
        \toprule
        \multicolumn{1}{c@{\hspace{0.6em}}|}{} & \multicolumn{7}{c|}{\textbf{Natural}} & \multicolumn{4}{c|}{\textbf{Specialized}} & \multicolumn{8}{c|}{\textbf{Structured}} & \multicolumn{3}{c}{\textbf{}}  \\ 
        ~ &\rotatebox{90}{CIFAR-100} &\rotatebox{90}{Caltech101} &\rotatebox{90}{DTD} &\rotatebox{90}{Flowers102} & \rotatebox{90}{Pets} &\rotatebox{90}{SVHN} &\rotatebox{90}{Sun397} &\rotatebox{90}{Camelyon} &\rotatebox{90}{EuroSAT} &\rotatebox{90}{Resisc45} &\rotatebox{90}{Retionpathy} &\rotatebox{90}{Clevr-Count} &\rotatebox{90}{Clevr-Dist} &\rotatebox{90}{DMLab} &\rotatebox{90}{KITTI-Dist} &\rotatebox{90}{dSpr-Loc} &\rotatebox{90}{dSpr-Ori} &\rotatebox{90}{sNORB-Azim} & \rotatebox{90}{sNORB-Elev} &\rotatebox{90}{Param. (M)} & \rotatebox{90}{FLOPs (G)} & \rotatebox{90}{Average}  \\ 
        \hline
        \multicolumn{23}{c}{\textit{Traditional methods}} \\ 
        Full tuning
        & 68.9 & 87.7 & 64.3 & 97.2 & 86.9 & 87.4 & 38.8 
        & 79.7 & 95.7 & 84.2 & 73.9 
        & 56.3 & 58.6 & 41.7 & 65.5 & 57.5 & 46.7 & 25.7 & 29.1 
        & 85.8 & 16.8 & 68.9 \\ 
        Linear probing
        & 63.4 & 85.0 & 63.2 & 97.0 & 86.3 & 36.6 & 51.0 
        & 78.5 & 87.5 & 68.6 & 74.0
        & 34.3 & 30.6 & 33.2 & 55.4 & 12.5 & 20.0 & 9.6 & 19.2 
        & 0.04 & 16.8 & 57.6 \\ 
        \hline
        \multicolumn{23}{c}{\textit{Parameter-efficient tuning methods}} \\ 
        Adapter~\cite{houlsby2019adapter}
        & 69.2 & 90.1 & 68.0 & 98.8 & 89.9 & 82.8 & 54.3 
        & 84.0 & 94.9 & 81.9 & 75.5
        & 80.9 & 65.3 & 48.6 & 78.3 & 74.8 & 48.5 & 29.9 & 41.6
        & 0.16 & 16.8 & 73.8 \\ 
        LoRA~\cite{hu2022lora}
        & 67.1 & 91.4 & 69.4 & 98.8 & 90.4 & 85.3 & 54.0
        & 84.9 & 95.3 & 84.4 & 73.6
        & 82.9 & 69.2 & 49.8 & 78.5 & 75.7 & 47.1 & 31.0 & 44.0
        & 0.29 & 16.8 & 74.6 \\ 
        \hline
        \multicolumn{23}{c}{\textit{Inference-efficient tuning methods}} \\ 
        $\text{DyT}_{0.7}$~\cite{zhao2024dyntuning}
        & \textbf{71.3} & \textbf{94.9} & 70.6 & 98.9 & 90.7 & 86.5 & \textbf{54.5} 
        & \textbf{87.8} & \textbf{96.2} & 85.5 & \textbf{76.7}
        & 82.7 & 61.4 & \textbf{53.1} & 82.0 & \textbf{85.7} & \textbf{53.8} & \textbf{34.6} & 42.5
        & 0.20 & 12.3 & \textbf{76.5} \\
         $\text{DeeBERT}_{0.7}$~\cite{xin2020deebert}
        & 62.9 & 90.8 & 69.9 & 98.8 & 86.3 & 81.8 & 53.4 
        & 81.2 & 94.5 & 84.8 & 72.2
        & 75.7 & 62.8 & 46.6 & 75.9 & 68.2 & 43.8 & 26.4 & 35.9
        & 0.46 & 11.6 & 71.8 \\ 
        $\text{Dyn-Adapter}_{0.7}$~\cite{zhang2024dynadapter}
        & 67.9 & 90.5 & 70.4 & 99.1 & 89.8 & 86.4 & 53.6
        & 86.3 & 95.7 & 84.3 & 75.1
        & 81.6 & 67.8 & 50.2 & \textbf{82.1} & 79.1 & 47.0 & 31.6 & 39.6
        & 0.46 & 11.6 & 75.0 \\ 
        $\text{DMPO}_{0.7}$ (Ours)
        & 69.2 & 92.5 & \textbf{71.3} & \textbf{99.4} & \textbf{90.9} & \textbf{89.4} & 54.3
        & 85.1 & \textbf{96.2} & \textbf{86.0} & 76.0
        & \textbf{83.6} & \textbf{70.2} & 51.3 & 80.9 & 81.7 & 48.0 & \textbf{34.6} & \textbf{42.8}
        & 0.87 & 11.6 & 76.2 \\
        \hline
        $\text{DyT}_{0.4}$~\cite{zhao2024dyntuning}
        & 63.4 & 90.8 & 64.5 & 96.8 & 86.6 & 84.5 & 51.5 
        & 83.4 & 92.6 & 80.4 & 74.1
        & 77.3 & 61.5 & \color{blue}\textbf{52.2} & 77.5 & \color{blue}\textbf{83.2} & \color{blue}\textbf{52.0} & 31.9 & 36.6
        & 0.20 & 6.7 & 72.8 \\
        $\text{DeeBERT}_{0.3}$~\cite{xin2020deebert}
        & 24.3 & 66.8 & 47.2 & 76.6 & 36.9 & 44.8 & 33.3 
        & 78.6 & 91.8 & 71.0 & 75.0
        & 51.7 & 43.3 & 34.2 & 60.0 & 27.9 & 23.7 & 14.3 & 24.6
        & 0.46 & 5.2 & 53.7 
        \\
        $\text{Dyn-Adapter}_{0.3}$~\cite{zhang2024dynadapter}
        & 66.8 & 90.1 & 69.3 & 99.0 & 89.4 & 84.6 & 53.7
        & 81.9 & 95.2 & 84.4 & \color{blue}\textbf{75.2}
        & 72.4 & 64.9 & 44.0 & 77.0 & 63.4 & 38.4 & 28.2 & 33.6
        & 0.46 & 5.2 & 72.0 \\ 
        $\text{DMPO}_{0.3}$ (Ours)
        & \color{blue}\textbf{68.6} & \color{blue}\textbf{92.4} & \color{blue}\textbf{71.2} & \color{blue}\textbf{99.4} & \color{blue}\textbf{90.2} & \color{blue}\textbf{89.4} & \color{blue}\textbf{54.3}
        & \color{blue}\textbf{85.0} & \color{blue}\textbf{96.2} & \color{blue}\textbf{86.1} & \color{blue}\textbf{75.2}
        & \color{blue}\textbf{83.2} & \color{blue}\textbf{70.2} & 51.0 & \color{blue}\textbf{80.8} & 81.6 & 47.6 & \color{blue}\textbf{34.8} & \color{blue}\textbf{44.1}
        & 0.87 & 5.2 & \color{blue}\textbf{76.0} \\ 
        \specialrule{.1em}{0em}{0em}
    \end{tabular}
    \setlength{\abovecaptionskip}{4pt} 
    \setlength{\belowcaptionskip}{5pt}
    \caption{Results on VTAB-1K benchmark. ``Param. (M)" denotes the number of trainable parameters. ``FLOPs (G)" is the average inference FLOPs across all datasets. ``Average" indicates the average accuracy of three groups.}
    \label{tab:vtab-1k}
\end{table*}

Intuitively, loss of earlier stage can serve as a control factor for subsequent stage. Specifically, a small loss of early stage indicates high discriminative ability, corresponding to reliable classification in early stage. On the contrary, a large loss suggests that the features of earlier stage account for imprecise classification and tend to be low-level and fundamental cues for subsequent stage. As such, we can use value of $\ell_{i-1}$ to regularize loss $\ell_{i}$ of $i$-th stage for each training sample, and we have
\begin{equation}
\setlength{\abovedisplayskip}{3pt}
\setlength{\belowdisplayskip}{3pt}
\hat{\ell}_{i} = \sigma(\ell_{i-1})\cdot{\ell_{i}}(\hat{Y}_{i},Y),  
\label{eq:beta}
\end{equation}
where $\sigma$ is an activation function to control the magnitude of influence, and \texttt{Sigmoid} function is used at default. For the first stage, we simply set $\hat{\ell}_{1} = \ell_1$.
As such, all factors involved in DMPO of \cref{eq:loss_total} are thoroughly elaborated. 

\vspace{-2mm}
\section{Experiments}\label{sec:Experiments}

\subsection{Experiment Settings}\label{sec:implementation}
\paragraph{Implementation Details.}
By default, we conduct experiments using ViT-B/16~\cite{dosovitskiy2020vit}, which is pre-trained on ImageNet-21K~\cite{deng2009imagenet} with supervision. We set the number of stages to $S=4$, with each stage containing $L=3$ blocks, and insert predictors uniformly. We select LoRA~\cite{hu2022lora} as the fine-tuning module, where the bottleneck dimension $d$ is set to 8. The start values of $\alpha$ are set to 0.01, 0.01, 1.0, and 2.0, respectively, while end values are set to 1.5, 1.0, 0.1, and 0.1. The weights at intermediate epochs are linearly interpolated between start and end values.
We set all the initial $\beta$ values to 1, which are adaptively updated during training. For all the tasks, we use top-1 accuracy as the primary evaluation metric, and FLOPs refers to inference FLOPs.
The numerical subscript indicates the FLOPs levels during inference for the method (\eg, $\text{DMPO}_{0.3}$ denotes DMPO with FLOPs controlled to 30\%).
The best results for low FLOPs and high FLOPs are highlighted in ${\color{blue}\textbf{blue bold}}$ and $\textbf{black bold}$, respectively.

\vspace{-1.5em}
\paragraph{Counterparts.}
We categorize the counterparts into three groups: (i) Traditional methods including full tuning all parameters and linear probing by only fine-tuning the classification head; 
(ii) PEFT methods including Adapter~\cite{houlsby2019adapter}, and LoRA~\cite{hu2022lora}; (iii) Inference-efficient methods including DyT~\cite{zhao2024dyntuning} with token pruning~\cite{zhao2024dyntuning}, and two early exiting methods DeeBERT~\cite{xin2020deebert} and Dyn-Adapter~\cite{zhang2024dynadapter}. 
Since DyT performs token pruning in the MLP, its inference can only be reduced to a minimum of 40\% FLOPs. Therefore, we compare our DMPO at 30\% FLOPs with DyT at 40\% FLOPs. DeeBERT is re-implemented by us since it is originally proposed for natural language processing. We follow the original paper for the adapters in Dyn-Adapter, using LoRA unless otherwise specified.

\begin{table*}[t]
    \centering
    \small
    \setlength{\tabcolsep}{4pt}
    \begin{tabular}{l|cc|cccccc|c}
    \toprule
    \textbf{Method} & \makecell{\textbf{Param.}\\\textbf{(M)}} & \makecell{\textbf{FLOPs}\\\textbf{(G)}} & \textbf{CIFAR-100} & \makecell{\textbf{CUB-200}\\\textbf{-2011}} & \textbf{NABirds} & \makecell{\textbf{Oxford}\\\textbf{Flowers}} &  \makecell{\textbf{Stanford}\\\textbf{Cars}} & \makecell{\textbf{Stanford}\\\textbf{Dogs}} &
    \textbf{Average}\\
    \midrule
    Linear probing
    & 0.17 & 16.76 & 88.7 & 85.3 & 75.9 & 97.9 & 51.3 & 86.2 & 80.88 \\
    Adapter~\cite{houlsby2019adapter} 
    & 0.56 & 16.79 & 93.3 & 87.1 & 84.3 & 98.5 & 68.6 & 89.8 & 86.93 \\
    LoRA~\cite{hu2022lora}
    & 0.56 & 16.76 & 93.7 & 88.7 & 84.9 & 99.3 & 89.7 & 88.4 & 90.78 \\
    \midrule
    $\text{DyT}_{0.7}$~\cite{zhao2024dyntuning}
    & 0.33 & 12.26 & 91.6 & 84.9 & 80.1 & \textbf{99.1} & 77.9 & \textbf{87.9} & 86.92 \\
    $\text{DeeBERT}_{0.7}$~\cite{xin2020deebert}
    & 0.95 & 11.61 & 86.1 & 67.5 & 70.2 & 98.2 & 72.6 & 68.6 & 77.20 \\
    $\text{Dyn-Adapter}_{0.7}$~\cite{zhang2024dynadapter}
    & 0.95 & 11.61 & 91.6 & 82.4 & 80.4 & 98.7 & 82.2 & 85.5 & 86.80 \\
    $\text{DMPO}_{0.7}$ (Ours)
    & 1.73 & 11.62 & \textbf{92.6} & \textbf{87.0} & \textbf{81.8} & \textbf{99.1} & \textbf{83.6} & 87.3 & \textbf{88.57} \\
    \hline
    $\text{DyT}_{0.4}$~\cite{zhao2024dyntuning}
    & 0.33 & 6.71 & 87.6 & 80.9 & 75.0 & 94.7 & 76.8 & 80.6 & $\text{82.60}_{(4.32)}$ \\
    $\text{DeeBERT}_{0.3}$~\cite{xin2020deebert}
    & 0.95 & 5.17 & 46.2 & 14.8 & 16.9 & 83.4 & 19.3 & 20.5 & $\text{33.52}_{(43.68)}$ \\
    $\text{Dyn-Adapter}_{0.3}$~\cite{zhang2024dynadapter}
    & 0.95 & 5.17 & 78.5 & 81.8 & 77.0 & 98.8 & 81.0 & 72.9 & $\text{81.67}_{(5.13)}$ \\
    $\text{DMPO}_{0.3}$ (Ours)
    & 1.73 & 5.19 & \color{blue}\textbf{92.3} & \color{blue}\textbf{86.1} & \color{blue}\textbf{80.8} & \color{blue}\textbf{99.1} & \color{blue}\textbf{83.7} & \color{blue}\textbf{86.5} & \color{blue}\textbf{$\text{88.08}_{(0.49)}$} \\
    \bottomrule
    \end{tabular}
    \setlength{\abovecaptionskip}{4pt} 
    \setlength{\belowcaptionskip}{4pt}
    \caption{Comparison of various methods on CIFAR-100 and five FGVC datasets. Note that the numbers in the subscripted parentheses indicate the decrease relative to the high FLOPs results.}
    \label{tab:fgvc}
\end{table*}

\vspace{-0.5em}
\subsection{Experiments on Downstream Datasets}
\paragraph{Evaluation by Utilizing VTAB-1K as Downstream Data.}
To evaluate the transfer learning performance of our approach, we conduct experiments on VTAB-1K~\cite{zhai2019vtab} benchmark. VTAB-1K is a collection of 19 diverse visual classification tasks, grouped as \textit{Natural}, \textit{Specialized}, and \textit{Structured}. Each task contains 1,000 labeled training images.
The results are shown in \cref{tab:vtab-1k}. With only 5.2 GFLOPs, approximately 30\% of the computational cost in original ViT-B/16, our DMPO achieves an accuracy of 76.0\%, outperforming PEFT methods and most inference-efficient fine-tuning methods. Notably, while our DMPO falls slightly short of the token pruning method DyT at around 70\% FLOPs, DyT’s accuracy drops significantly (\ie, by 3.7\%) when reduced to 40\% FLOPs. In contrast, our method exhibits only a minimal decrease of 0.2\%. Furthermore, in comparison to early exiting methods such as DeeBERT and Dyn-Adapter, our approach achieves higher accuracy at just 5.2 GFLOPs, surpassing their performance even at 11.6 GFLOPs (\ie, 70\% of the original ViT-B/16).

\paragraph{Evaluation by Utilizing More Downstream Data.}
To further demonstrate the performance and efficiency of the proposed method, we extend the experiments to complete image datasets, including CIFAR-100~\cite{krizhevsky2009cifar} and five FGVC datasets (\ie, CUB-200-2011~\cite{wah2011cub}, NABirds~\cite{van2015nabirds}, Oxford Flowers~\cite{nilsback2008flowers}, Stanford Dogs~\cite{khosla2011dogs}, and Stanford Cars~\cite{gebru2017cars}). 
As shown in \cref{tab:fgvc}, our DMPO demonstrates strong performance with more training data, achieving an average accuracy of 88.08\% at 5.19 GFLOPs and 88.57\% at 11.62 GFLOPs. Notably, at approximately 30\% FLOPs, DMPO outperforms previous inference-efficient fine-tuning methods, particularly DyT, evaluated at 70\% FLOPs. Furthermore, with a 30\% reduction in FLOPs, DyT exhibits a 4.32\% drop in accuracy, while Dyn-Adapter decreases by 5.13\% with a 40\% FLOPs reduction. In contrast, our DMPO demonstrates only a minimal reduction of 0.49\%, highlighting its robustness and stability.

\begin{table}
    \centering
    \small
    \begin{tabular}{l cccc}
    \toprule
    \multirow{2}{*}{\textbf{Method}} & \multicolumn{2}{c}{\textbf{SUP}} & \multicolumn{2}{c}{\textbf{SSL}} \\ 
    \cmidrule(lr){2-3} \cmidrule(lr){4-5}
    & DeiT-S & ViT-L & MoCo v3 & MAE \\
    \midrule
    $\text{DyT}_{0.7}$~\cite{zhao2024dyntuning} 
    & 82.02 & 90.41 & 79.89 & 75.62 \\
    $\text{Dyn-Adapter}_{0.7}$~\cite{zhang2024dynadapter} 
    & 79.22 & 89.94 & 81.13 & 74.31 \\
    $\text{DMPO}_{0.7}$ (Ours)
    & \textbf{82.33} & \textbf{90.77} & \textbf{82.91} & \textbf{76.58}  \\
    \hline
    $\text{DyT}_{0.4}$~\cite{zhao2024dyntuning} 
    & 73.94 & 85.59 & 76.49 & 63.93 \\
    $\text{Dyn-Adapter}_{0.3}$~\cite{zhang2024dynadapter} 
    & 62.47 & 89.15 & 72.27 & 66.63 \\
    $\text{DMPO}_{0.3}$ (Ours)
    & \color{blue}\textbf{82.20} & \color{blue}\textbf{90.64} & \color{blue}\textbf{82.49} & \color{blue}\textbf{76.33}  \\
    \bottomrule
    \end{tabular}
    \caption{Comparison with various pre-trained models, reporting the average accuracy of CIFAR-100 and FGVC datasets.}
    \label{tab:model_variant}
\end{table}
\vspace{-2mm}

\vspace{-1em}
\paragraph{Evaluation with Different Pre-trained Models.}
Besides, we explore the performance of DMPO across various model choices. Specifically, we consider two distinct categories: (i) Different model scales such as DeiT-S~\cite{touvron2021deit} and ViT-L/16~\cite{dosovitskiy2020vit}, pre-trained with supervised learning on ImageNet-1K and ImageNet-21K, respectively; (ii) ViT-B/16 pre-trained with self-supervised learning, \ie, MoCo v3~\cite{chen2021mocov3} and MAE~\cite{he2022mae}. 
According to \cref{tab:model_variant}, our DMPO demonstrates robust performance with minimal accuracy degradation across different pre-trained models, in contrast to the substantial performance degradation observed in DyT and Dyn-Adapter under significantly reduced FLOPs. Furthermore, DMPO consistently outperforms both Dyn-Adapter and DyT across all model variants at 30\% and 70\% FLOPs. Notably, our approach achieves higher accuracy at 30\% FLOPs than theirs at 70\% FLOPs.

\vspace{-1em}
\subsection{Comparisons on Different Scenarios}
\vspace{-1.5mm}
\begin{figure}
    \centering
    \includegraphics[width=0.35\textwidth]{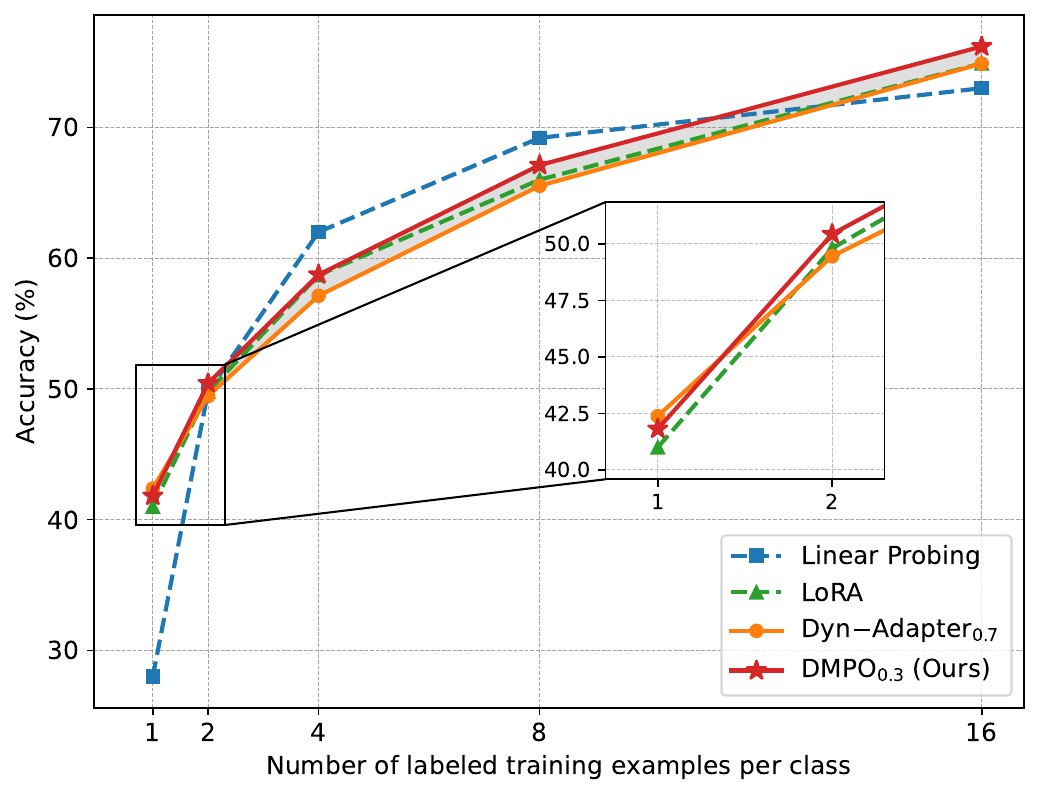}
    \setlength{\abovecaptionskip}{4pt} 
    \setlength{\belowcaptionskip}{4pt}
    \caption{Results of few-shot learning on five fine-grained visual recognition datasets. We employ RepAdapter as the adapters within Dyn-Adapter, following its original paper.}
    \label{fig:few_shot}
\end{figure}

\paragraph{Few-shot Learning.}
Following the settings in~\cite{zhang2022noah}, we conduct few-shot learning experiments on five fine-grained datasets (\ie, Food-101~\cite{bossard2014food}, Stanford Cars, Oxford Flowers, FGVCAircraft~\cite{maji2013aircraft} and OxfordPets~\cite{parkhi2012pets}).
As illustrated in \cref{fig:few_shot}, compared to Dyn-Adapter at 70\% FLOPs, our DMPO achieves better performance under larger-shot settings (\ie, $>$ 1-shot) at 30\% FLOPs. Furthermore, compared to the original LoRA, our DMPO consistently demonstrates superior performance across various shots at 30\% FLOPs. In the 16-shot setting, our method attains the best accuracy of 76.2\%, achieving notable advancements in accuracy and computational efficiency.

\begin{table}
    \centering
    \small
    \setlength{\tabcolsep}{2pt}
    \begin{tabular}{l cccccccc}
    \toprule
    \multirow{2}{*}{\textbf{Method}} & \textbf{Source} & \multicolumn{4}{c}{\textbf{Target}} \\ 
    \cmidrule(lr){2-2} \cmidrule(lr){3-6}
    & ImageNet & -V2 & -Sketch & -A & -R \\
    \midrule
    Adapter~\cite{houlsby2019adapter} 
    & 70.5 & 59.1 & 16.4 & 5.5 & 22.1 \\
    LoRA~\cite{hu2022lora} 
    & 70.8 & 59.3 & 20.0 & 6.9 & 23.3 \\
    \midrule
    $\text{DyT}_{0.7}$~\cite{zhao2024dyntuning} 
    & 75.4 & \textbf{63.6} & \textbf{25.7} & 11.4 & 26.6 \\
    $\text{Dyn-Adapter}_{0.75}$~\cite{zhang2024dynadapter} 
    & 71.0 & 59.3 & 20.7 & 7.3 & 22.5 \\
    $\text{DMPO}_{0.7}$ (Ours)
    & \textbf{75.9} & \textbf{63.6} & 21.5 & \textbf{12.5} & \textbf{26.7} \\
    \hline
    $\text{DyT}_{0.4}$~\cite{zhao2024dyntuning} 
    & 64.7 & 53.6 & 18.1 & 5.2 & 20.0 \\
    $\text{DMPO}_{0.3}$ (Ours)
    & \color{blue}\textbf{75.9} & \color{blue}\textbf{64.2} & \color{blue}\textbf{20.5} & \color{blue}\textbf{12.2} & \color{blue}\textbf{25.3} \\
    \bottomrule
    \end{tabular}
    \setlength{\abovecaptionskip}{3pt} 
    \setlength{\belowcaptionskip}{3pt}
    \caption{Results on domain generalization. We use the 75\% FLOPs results from the Dyn-Adapter paper. Since Dyn-Adapter performs poorly at 30\% FLOPs, we do not present its results here.}
    \label{tab:domain_generalization}
\end{table}

\vspace{-1.5em}
\paragraph{Domain Generalization.}
Following previous study~\cite{zhang2024dynadapter}, we fine-tune the model on ImageNet with 16-shot setting and evaluate its domain generalization ability by directly adapting to four variants of ImageNet (\ie, ImageNet V2~\cite{recht2019imagenetv2}, ImageNet-Sketch~\cite{wang2019imagenetsketch}, ImageNet-A~\cite{hendrycks2021imageneta} and ImageNet-R~\cite{hendrycks2021imagenetr}). 
As shown in \cref{tab:domain_generalization}, our DMPO outperforms Dyn-Adapter while using only 30\% of the computation, achieving improvements of 4.9\%, 4.9\%, and 2.8\% on ImageNet-V2, -A, and -R, respectively. Compared to the token pruning method DyT at 70\% FLOPs, our DMPO achieves comparable accuracy with 40\% fewer FLOPs. 

\subsection{Ablation Studies}
To validate the effectiveness of our DMPO, we conduct ablation studies on CIFAR-100 and five FGVC datasets.
\vspace{-1.0em}
\paragraph{Component Analysis.} 
Our DMPO consists of two components, \ie, Multi-Predictor Architecture (MPA) and Decoupled Optimization (DO). \cref{tab:component} illustrates the impact of each component, where $\checkmark$ indicates that the corresponding component is used, while \ding{55} denotes it is not used. From \cref{tab:component}, we can see that both MPA and DO improve performance compared to Dyn-Adapter~\cite{zhang2024dynadapter}.

\begin{table}[t]
  \centering
  \small
  \setlength{\tabcolsep}{1.3em}
    \begin{tabular}{cc|cc}
    \toprule
    \multicolumn{2}{c|}{\textbf{Component}} & \multicolumn{2}{c}{\textbf{Average Acc (\%)}} \\
    \midrule
    \multicolumn{1}{c}{MPA} & \multicolumn{1}{c|}{DO} & 30\% FLOPs & 70\% FLOPs \\
    \midrule
    \rowcolor{yellow!20}
    \ding{55}  & \ding{55}  & 81.67 & 86.80 \\
    \checkmark & \ding{55}  & 85.16 & 86.98 \\
    \ding{55}  & \checkmark & 84.41 & 86.96 \\
    \rowcolor[gray]{0.9} 
    \checkmark & \checkmark & \color{blue}\textbf{88.08} & \textbf{88.57} \\
    \bottomrule
    \end{tabular}%
\setlength{\abovecaptionskip}{3pt} 
\setlength{\belowcaptionskip}{3pt}
\caption{Average results with various components. Note that \colorbox{Yellow!20}{yellow} represents Dyn-Adapter, while \colorbox{gray!20}{gray} represents DMPO.}
\label{tab:component}
\end{table}
\vspace{-2mm}

\vspace{-0.5em}
\paragraph{Different Loss Weight Allocation Strategies.} 
To validate the effectiveness of our loss weight allocation method discussed in \cref{sec:alpha}, we conduct ablation studies on various allocation strategies. As shown in \cref{tab:alpha_change}, allocation strategies that maintain consistently high weights for either early or deep stages fail to achieve strong performance at both low and high FLOPs levels. Moreover, initially assigning a large weight to early stages, followed by a reduction, yields good performance at low FLOPs but leads to decreased performance at higher FLOPs.

\begin{table}[t]
  \centering
  \small
  \setlength{\tabcolsep}{1.2em}
    \begin{tabular}{cc|cc}
    \toprule
    \multicolumn{2}{c|}{\textbf{Training Phase}} & \multicolumn{2}{c}{\textbf{Average Acc} (\%)} \\
    \midrule
    \multicolumn{1}{c}{Initial} & \multicolumn{1}{c|}{Latter} & 30\% FLOPs & 70\% FLOPs \\
    \midrule
    $<$ & $<$    & 85.03 & 85.24 \\
    $>$ & $>$    & 83.86 & 85.80 \\
    $>$ & $<$    & 86.43 & 86.62 \\
    \rowcolor[gray]{0.9} $<$  & $>$    & \color{blue}\textbf{88.08} & \textbf{88.57} \\
    \bottomrule
    \end{tabular}%
\setlength{\abovecaptionskip}{3pt} 
\setlength{\belowcaptionskip}{3pt}
\caption{Average results for loss weight allocation. Note that $>$ indicates $\alpha_{\text{early}} > \alpha_{\text{deep}}$, $<$ indicates $\alpha_{\text{early}} < \alpha_{\text{deep}}$.}
\label{tab:alpha_change}
\end{table}
\vspace{-2mm}

\vspace{-2mm}
\paragraph{Different Activation Functions.}
As shown in \cref{eq:beta}, we experiment with different activation functions for the multi-stage loss. As shown in \cref{tab:beta_change}, \ding{55} denotes that the method for reducing inter-stage influence, discussed in \cref{sec:reduce_inf}, is not applied. \cref{tab:beta_change} demonstrates that reducing inter-stage influence improves the performance at 70\% FLOPs, indirectly suggesting that this approach can mitigate the impact of early stages on deep stages during the learning of discriminative features.

\begin{table}[t]
  \centering
  \small
  \setlength{\tabcolsep}{1.1em}
    \begin{tabular}{l|cc}
    \toprule
    \multirow{2}{*}{\textbf{Activation}} & \multicolumn{2}{c}{\textbf{Average Acc} (\%)} \\
\cmidrule{2-3}          & 30\% FLOPs  & 70\% FLOPs \\
    \midrule
    \ding{55} & 87.85 & 88.02  \\
    $\ell_i$ & 83.81 & 85.53  \\
    $\texttt{Tanh}(\ell_i)$ & 80.69 & 84.35 \\
    $\texttt{Softmax}(\ell_i)$ & 74.57 & 75.74  \\
    \rowcolor[gray]{0.9} $\texttt{Sigmoid}(\ell_i)$ & \color{blue}\textbf{88.08} & \textbf{88.57}  \\
    \midrule    
    \end{tabular}%
      \setlength{\abovecaptionskip}{3pt} 
      \setlength{\belowcaptionskip}{3pt}
      \caption{Average results of different activation functions.}
  \label{tab:beta_change}%
\end{table}%


\vspace{-2mm}
\section{Conclusion}
\vspace{-2mm}
In this paper, we propose an inference-efficient approach based on early exiting to adapt pre-trained large-scale models for downstream applications. Our proposed DMPO method incorporates a modified multi-predictor architecture along with a decoupled optimization algorithm. In terms of the architecture, lightweight bypass modules and high-order statistics-based predictors are employed to decouple representative ability and discriminative ability in early stages. Subsequently, through decoupled optimization, the model can acquire discriminative ability, which can be driven towards earlier stages as much as possible. Across downstream datasets and model backbones, our DMPO significantly reduces inference costs while maintaining superior accuracy compared to the counterparts. In future work, our DMPO can be extended to achieve inference-efficient models from pre-trained large-scale models in natural language processing and multi-modal domains.

\section{Acknowledgment}
This work was supported in part by the National Natural Science Foundation of China under Grant 62222608, 62276186, U23B2049, 62172127 and 62436002.

\clearpage
{
    \small
    \bibliographystyle{ieeenat_fullname}
    \bibliography{main}
}
\maketitlesupplementary

\renewcommand{\thetable}{S\arabic{table}}
\renewcommand{\thefigure}{S\arabic{figure}}
\renewcommand{\thesection}{S\arabic{section}}
\renewcommand{\theequation}{S\arabic{equation}}
\renewcommand{\thefootnote}

\setcounter{page}{1}
\setcounter{section}{0}
\setcounter{table}{0}
\setcounter{equation}{0}
\setcounter{figure}{0}

In the supplementary material, we conduct more experiments to further investigate the effectiveness of our DMPO.
Specifically, we first provide a detailed explanation of the experiments shown in \cref{fig:similarity} and \cref{fig:similarity_res}, and conduct the same experiments on five FGVC datasets. Subsequently, we evaluate the generalization ability of DMPO using different fine-tuning modules. Additionally, we perform experiments on a larger dataset (\ie, ImageNet-1K~\cite{deng2009imagenet}). Next, we conduct ablation studies on various components of DMPO and present an analysis of the DMPO training process and overhead. Finally, we visualize the classified images at different stages.

\section{Decoupling Challenge in Early Stages}
To investigate why early exiting networks cannot maintain the same high performance in both early and deep stages simultaneously, we design an experiment on CIFAR-100~\cite{krizhevsky2009cifar} dataset to calculate the cosine similarity between two variants of Dyn-Adapter~\cite{zhang2024dynadapter} and Original ViT.
In this experiment, to better investigate the influence of early stages on deep stages, both Dyn-Adapter and its variant are configured with $S=2$ stages, each consisting of $L=6$ blocks.

As shown in \cref{fig:similarity} and \cref{fig:similarity_res}, Dyn-Adapter adopts the loss weight allocation strategy proposed in~\cite{zhang2024dynadapter}, with the output feature of Stage 1 denoted as $\textbf{X}_\text{Dyn}$.
The output feature of Original ViT in the 6-th block is denoted as $\textbf{X}_\text{Ori}$, and we consider $\textbf{X}_\text{Ori}$ to be the low-level representative feature required by deeper stage.
Dyn-R indicates that based on Dyn-Adapter, the variation in $\textbf{X}_\text{Dyn}$ is restricted using $\lVert\textbf{X}_\text{Ori}-\textbf{X}_\text{Dyn}\rVert_2$, resulting in a new feature representation, $\textbf{X}_\text{R}$.
By calculating the cosine similarity of $(\textbf{X}_\text{Ori}, \textbf{X}_\text{Dyn})$ and $(\textbf{X}_\text{Ori}, \textbf{X}_\text{R})$, we obtain the results shown in \cref{fig:similarity}.
As illustrated in \cref{fig:similarity_res}, ``Stage 1" indicates that all samples exit at Stage 1, whereas ``Stage 2" indicates that all samples exit at Stage 2.

As training progresses, $\textbf{X}_\text{Dyn}$, which is directly involved in classification learning, will contain more discriminative information than $\textbf{X}_\text{Ori}$, while $\textbf{X}_\text{R}$ will consistently maintain a fundamental representation highly similar to $\textbf{X}_\text{Ori}$ due to $\lVert\textbf{X}_\text{Ori}-\textbf{X}_\text{Dyn}\rVert_2$. From \cref{fig:similarity} and \cref{fig:similarity_res}, the performance of early stage declines when providing representative features instead of discriminative features. At the same time, the performance of deep stage shows the opposite trend. This indicates a conflict where early stage struggles to learn representative and discriminative features simultaneously. In light of the preceding analysis, we introduce our DMPO method to address this conflict.
Under similar experiment settings, as shown in \cref{fig:similarity} and \cref{fig:similarity_res}, the output features of early stages in our DMPO achieve a trade-off between fundamental and discriminative information, achieving the best results at both Stage 1 and Stage 2.

\begin{figure*}[t]
    \centering
    \includegraphics[width=1.0\textwidth]{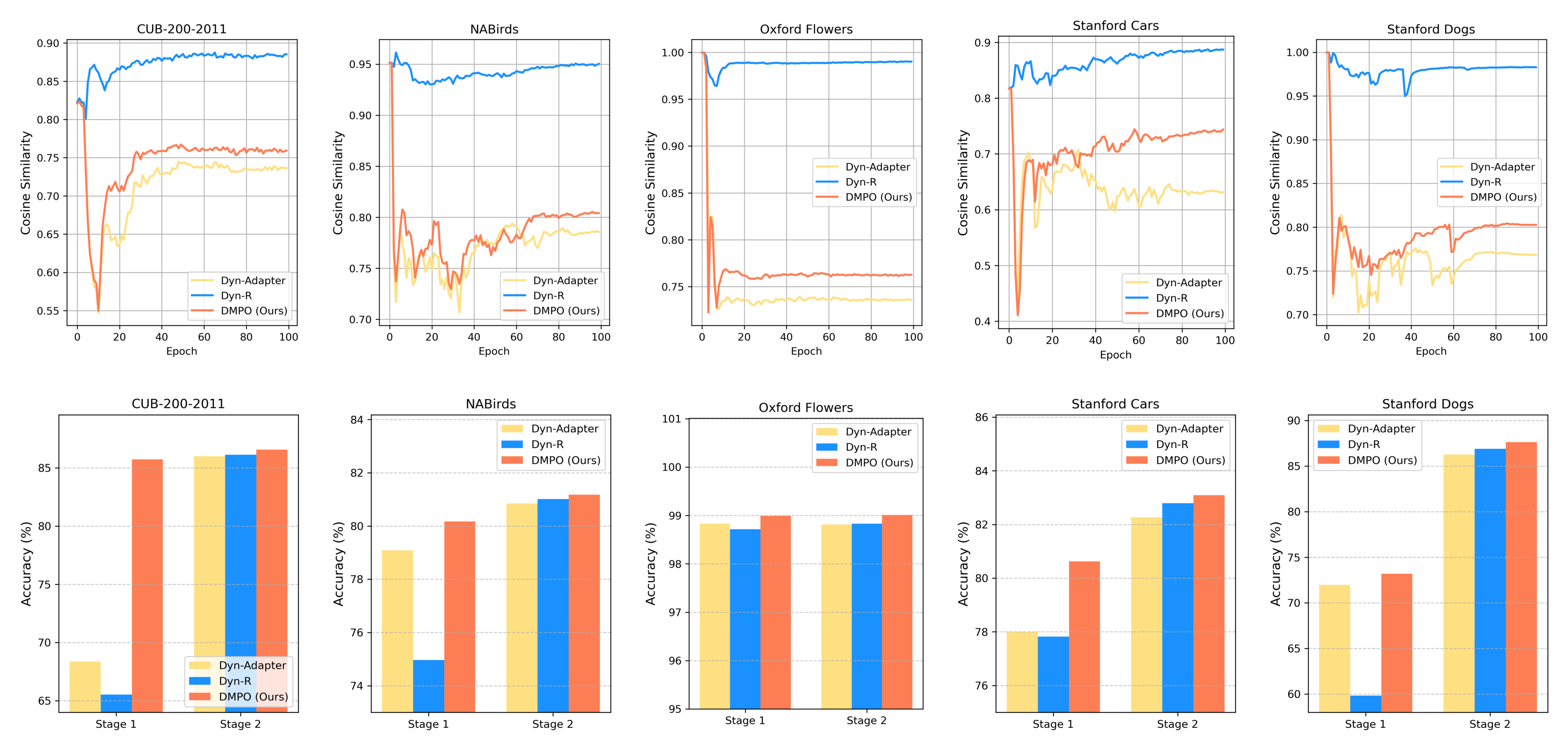}
    \caption{Comparison of cosine similarities (top row) among Dyn-Adapter, Dyn-R, DMPO, and Original ViT, and classification accuracies (bottom row) of Dyn-Adapter, Dyn-R, and DMPO at Stage 1 and Stage 2 on five FGVC datasets.}
    \label{fig:sim_other_datasets}
\end{figure*}

To thoroughly validate the unresolved decoupling conflict in early stages, we conduct the same experiments on five FGVC datasets. The results are presented in \cref{fig:sim_other_datasets}.

\section{Theoretical Analysis on Decoupled Optimization}
Analogous to Eq.(3) in DSN~\cite{lee2015deeply}, total loss of a model with multi-stage predictors $F(\bm{w})$ can be expressed as $F(\bm{w}) = P(\bm{w}) + Q(\bm{w})$, where $P(\bm{w})$ denotes final classification loss, and $Q(\bm{w})$ indicates sum loss of all preceding predictors. Based on Lemma 1 in DSN~\cite{lee2015deeply}, we know $F(\bm{w})$ and $P(\bm{w})$ share the same optimal weights $\bm{w}^{*}$, while $Q(\bm{w})$ may attain a different optimal $\bm{w}^{+}$. For our DMPO, we require both $\bm{w}^{*}$ and $\bm{w}^{+}$. Therefore, our two-phase optimization obtains the optimal solution $\bm{w}^{*}$ (representation) for $F(\bm{w})$ at first phase, and then optimize $Q(\bm{w})$ to achieve $\bm{w}^{+}$ (discrimination) in second phase while keeping $\bm{w}^{*}$ as much as possible, constrained by $\alpha_i$ and $\sigma(\ell_{i-1})$. The detailed theoretical analysis will be added in revision. 

\section{Performances on CNN Architecture}
We experiment with the backbone of ConvNeXt-Base to validate effect of Dyn-Adapter and our DMPO on CNN by tuning on CUB-200-2011, where full inference with LoRA tuning achieves a top-1 accuracy of 88.66\%. As illustrated in \cref{tab:cnn}, for 70\% FLOPs, DMPO clearly outperforms Dyn-Adapter. For 30\% FLOPs, DMPO significantly surpasses Dyn-Adapter. These results verify effectiveness of DMPO for CNNs.

\begin{table}[h]
 \centering
 \small
   \begin{tabular}{cccc}
   \toprule
    \multicolumn{1}{c}{\textbf{Method}} & \multicolumn{1}{c}{\textbf{Acc\textsubscript{0.3}}} & \multicolumn{1}{c}{\textbf{Acc\textsubscript{0.7}}} & \multicolumn{1}{c}{\textbf{Acc\textsubscript{1.0}}} \\
   \midrule
    Dyn-Adapter & 31.83 & 74.65 & 88.76 \\
    DMPO & 82.26 & 86.97 & 88.21 \\
   \bottomrule
   \end{tabular}%
   \setlength{\abovecaptionskip}{3pt} 
   \setlength{\belowcaptionskip}{3pt}
\caption{ ConvNeXt-Base as backbone on CUB-200-2011.}
\label{tab:cnn}
\end{table}

\vspace{-1em}
\section{Different Fine-tuning Modules}
In \cref{sec:Experiments}, we use LoRA~\cite{hu2022lora} as the fine-tuning module. To further validate the generalization ability of our DMPO, we conduct experiments using other fine-tuning modules (\ie, Adapter~\cite{houlsby2019adapter} and Repadapter~\cite{luo2023repadapter}). The results on VTAB-1K are presented in \cref{tab:sup-vtab-1k}, while those on CIFAR-100 and five FGVC datasets are shown in \cref{tab:sup-fgvc}.
As shown in \cref{tab:sup-vtab-1k}, although our DMPO exhibits a slight performance drop compared to Dyn-Adapter when using Repadapter as the fine-tuning module, it demonstrates better performance at low FLOPs and maintains greater stability across varying FLOPs. Furthermore, \cref{tab:sup-fgvc} indicates that when utilizing more downstream data, our DMPO consistently achieves the best performance at both low and high FLOPs.

\begin{table*}[t!]
    \centering
    \small
    \setlength{\tabcolsep}{1.5pt}
    \begin{tabular}{l@{\hspace{0.2em}}|ccccccc|cccc|cccccccc|ccc}
        \toprule
        \multicolumn{1}{c@{\hspace{0.6em}}|}{} & \multicolumn{7}{c|}{\textbf{Natural}} & \multicolumn{4}{c|}{\textbf{Specialized}} & \multicolumn{8}{c|}{\textbf{Structured}} & \multicolumn{3}{c}{\textbf{}}  \\ 
        ~ &\rotatebox{90}{CIFAR-100} &\rotatebox{90}{Caltech101} &\rotatebox{90}{DTD} &\rotatebox{90}{Flowers102} & \rotatebox{90}{Pets} &\rotatebox{90}{SVHN} &\rotatebox{90}{Sun397} &\rotatebox{90}{Camelyon} &\rotatebox{90}{EuroSAT} &\rotatebox{90}{Resisc45} &\rotatebox{90}{Retionpathy} &\rotatebox{90}{Clevr-Count} &\rotatebox{90}{Clevr-Dist} &\rotatebox{90}{DMLab} &\rotatebox{90}{KITTI-Dist} &\rotatebox{90}{dSpr-Loc} &\rotatebox{90}{dSpr-Ori} &\rotatebox{90}{sNORB-Azim} & \rotatebox{90}{sNORB-Elev} &\rotatebox{90}{Param. (M)} & \rotatebox{90}{FLOPs (G)} & \rotatebox{90}{Average}  \\ 
        \hline
        \rowcolor[gray]{0.9}\multicolumn{23}{c}{\textit{LoRA}~\cite{hu2022lora}} \\ 
        $\text{Dyn-Adapter}_{0.7}$~\cite{zhang2024dynadapter}
        & 67.9 & 90.5 & 70.4 & 99.1 & 89.8 & 86.4 & 53.6
        & \textbf{86.3} & 95.7 & 84.3 & 75.1
        & 81.6 & 67.8 & 50.2 & \textbf{82.1} & 79.1 & 47.0 & 31.6 & 39.6
        & 0.46 & 11.6 & 75.0 \\ 
        $\text{DMPO}_{0.7}$ (Ours)
        & \textbf{69.2} & \textbf{92.5} & \textbf{71.3} & \textbf{99.4} & \textbf{90.9} & \textbf{89.4} & \textbf{54.3}
        & 85.1 & \textbf{96.2} & \textbf{86.0} & \textbf{76.0}
        & \textbf{83.6} & \textbf{70.2} & \textbf{51.3} & 80.9 & \textbf{81.7} & \textbf{48.0} & \textbf{34.6} & \textbf{42.8}
        & 0.87 & 11.6 & \textbf{ 76.2} \\
        $\text{Dyn-Adapter}_{0.3}$~\cite{zhang2024dynadapter}
        & 66.8 & 90.1 & 69.3 & 99.0 & 89.4 & 84.6 & 53.7
        & 81.9 & 95.2 & 84.4 & \color{blue}\textbf{75.2}
        & 72.4 & 64.9 & 44.0 & 77.0 & 63.4 & 38.4 & 28.2 & 33.6
        & 0.46 & 5.2 & 72.0 \\ 
        $\text{DMPO}_{0.3}$ (Ours)
        & \color{blue}\textbf{68.6} & \color{blue}\textbf{92.4} & \color{blue}\textbf{71.2} & \color{blue}\textbf{99.4} & \color{blue}\textbf{90.2} & \color{blue}\textbf{89.4} & \color{blue}\textbf{54.3}
        & \color{blue}\textbf{85.0} & \color{blue}\textbf{96.2} & \color{blue}\textbf{86.1} & \color{blue}\textbf{75.2}
        & \color{blue}\textbf{83.2} & \color{blue}\textbf{70.2} & \color{blue}\textbf{51.0} & \color{blue}\textbf{80.8} & \color{blue}\textbf{81.6} & \color{blue}\textbf{47.6} & \color{blue}\textbf{34.8} & \color{blue}\textbf{44.1}
        & 0.87 & 5.2 & \color{blue}\textbf{76.0} \\ 
        \hline
        \rowcolor[gray]{0.9}\multicolumn{23}{c}{\textit{Adapter}~\cite{houlsby2019adapter}} \\ 
        $\text{Dyn-Adapter}_{0.7}$~\cite{zhang2024dynadapter}
        & 68.3 & 91.1 & 67.6 & 98.9 & 89.5 & 85.7 & 54.3
        & 83.0 & 95.8 & 84.6 & 75.8
        & 80.4 & 64.8 & \textbf{48.5} & 78.5 & 76.0 & 49.4 & 30.0 & 40.0
        & 0.28 & 11.6 & 74.2 \\ 
        $\text{DMPO}_{0.7}$ (Ours)
        & \textbf{68.9} & \textbf{92.6} & \textbf{70.4} & \textbf{99.1} & \textbf{90.4} & \textbf{88.0} & \textbf{54.5}
        & \textbf{84.5} & \textbf{96.0} & \textbf{85.0} & \textbf{76.0}
        & \textbf{81.8} & \textbf{65.1} & 48.4 & \textbf{79.8} & \textbf{76.6} & \textbf{50.6} & \textbf{30.8} & \textbf{41.9}
        & 0.78 & 11.6 & \textbf{75.1} \\
        $\text{Dyn-Adapter}_{0.3}$~\cite{zhang2024dynadapter}
        & 62.9 & 90.8 & 62.9 & 98.4 & 81.0 & 81.0 & 53.6
        & 80.0 & 94.8 & 82.1 & 75.4
        & 65.4 & 60.9 & 39.2 & 69.2 & 37.7 & 32.4 & 19.7 & 30.9
        & 0.28 & 5.2 & 67.8 \\ 
        $\text{DMPO}_{0.3}$ (Ours)
        & \color{blue}\textbf{68.8} & \color{blue}\textbf{92.5} & \color{blue}\textbf{70.2} & \color{blue}\textbf{99.1} & \color{blue}\textbf{90.1} & \color{blue}\textbf{88.1} & \color{blue}\textbf{54.2}
        & \color{blue}\textbf{84.6} & \color{blue}\textbf{96.0} & \color{blue}\textbf{85.2} & \color{blue}\textbf{75.5}
        & \color{blue}\textbf{81.6} & \color{blue}\textbf{65.6} & \color{blue}\textbf{48.0} & \color{blue}\textbf{78.3} & \color{blue}\textbf{78.3} & \color{blue}\textbf{51.3} & \color{blue}\textbf{31.1} & \color{blue}\textbf{44.6}
        & 0.78 & 5.2 & \color{blue}\textbf{75.2} \\ 
        \hline
        \rowcolor[gray]{0.9}\multicolumn{23}{c}{\textit{Repadapter}~\cite{luo2023repadapter}} \\ 
        $\text{Dyn-Adapter}_{0.7}$~\cite{zhang2024dynadapter}
        & \textbf{71.8} & 92.6 & \textbf{71.7} & \textbf{99.1} & \textbf{90.6} & 90.8 & \textbf{54.3}
        & \textbf{85.8} & 95.9 & \textbf{86.4} & \textbf{76.1}
        & 80.3 & \textbf{68.9} & 49.9 & \textbf{81.9} & \textbf{82.3} & \textbf{50.3} & \textbf{36.8} & 41.1
        & 0.38 & 11.6 &\textbf{ 76.4} \\ 
        $\text{DMPO}_{0.7}$ (Ours)
        & 70.1 & \textbf{93.4} & \textbf{71.7} & \textbf{99.1} & 90.5 & \textbf{91.1} & 53.6
        & 85.3 & \textbf{96.1} & \textbf{86.4} & 75.9
        & \textbf{81.8} & \textbf{68.9} & \textbf{51.0} & 81.0 & 80.7 & 49.6 & 34.4 & \textbf{43.0}
        & 0.78 & 11.6 & 76.2 \\
        $\text{Dyn-Adapter}_{0.3}$~\cite{zhang2024dynadapter}
        & 69.0 & 92.5 & 68.0 & 98.9 & 88.1 & 86.4 & \color{blue}\textbf{53.7}
        & 81.9 & 95.5 & 85.2 & 74.8
        & 69.9 & 64.3 & 44.1 & 77.4 & 70.4 & 40.6 & 28.3 & 33.3
        & 0.38 & 5.2 & 72.4 \\ 
        $\text{DMPO}_{0.3}$ (Ours)
        & \color{blue}\textbf{69.8} & \color{blue}\textbf{93.4} & \color{blue}\textbf{71.4} & \color{blue}\textbf{99.1} & \color{blue}\textbf{90.0} & \color{blue}\textbf{91.1} & 53.6
        & \color{blue}\textbf{85.4} & \color{blue}\textbf{96.2} & \color{blue}\textbf{86.4} & \color{blue}\textbf{75.5}
        & \color{blue}\textbf{81.7} & \color{blue}\textbf{68.9} & \color{blue}\textbf{50.9} & \color{blue}\textbf{81.5} & \color{blue}\textbf{81.3} & \color{blue}\textbf{49.8} & \color{blue}\textbf{34.3} & \color{blue}\textbf{44.9}
        & 0.78 & 5.2 & \color{blue}\textbf{76.2} \\
        \specialrule{.1em}{0em}{0em}
    \end{tabular}
    \caption{Results of different fine-tuning modules on VTAB-1K benchmark.}
    \label{tab:sup-vtab-1k}
\end{table*}

\begin{table*}[t]
    \centering
    \small
    \setlength{\tabcolsep}{4pt}
    \begin{tabular}{l|cc|cccccc|c}
    \toprule
    \textbf{Method} & \makecell{\textbf{Param.}\\\textbf{(M)}} & \makecell{\textbf{FLOPs}\\\textbf{(G)}} & \textbf{CIFAR-100} & \makecell{\textbf{CUB-200}\\\textbf{-2011}} & \textbf{NABirds} & \makecell{\textbf{Oxford}\\\textbf{Flowers}} &  \makecell{\textbf{Stanford}\\\textbf{Cars}} & \makecell{\textbf{Stanford}\\\textbf{Dogs}} &
    \textbf{Average}\\
    \midrule
    \rowcolor[gray]{0.9}\multicolumn{10}{c}{\textit{LoRA}~\cite{hu2022lora}} \\ 
    $\text{Dyn-Adapter}_{0.7}$~\cite{zhang2024dynadapter}
    & 0.95 & 11.61 & 91.6 & 82.4 & 80.4 & 98.7 & 82.2 & 85.5 & 86.80 \\
    $\text{DMPO}_{0.7}$ (Ours)
    & 1.73 & 11.62 & \textbf{92.6} & \textbf{87.0} & \textbf{81.8} & \textbf{99.1} & \textbf{83.6} & \textbf{87.3} & \textbf{88.57} \\
    $\text{Dyn-Adapter}_{0.3}$~\cite{zhang2024dynadapter}
    & 0.95 & 5.17 & 78.5 & 81.8 & 77.0 & 98.8 & 81.0 & 72.9 & $\text{81.67}$ \\
    $\text{DMPO}_{0.3}$ (Ours)
    & 1.73 & 5.19 & \color{blue}\textbf{92.3} & \color{blue}\textbf{86.1} & \color{blue}\textbf{80.8} & \color{blue}\textbf{99.1} & \color{blue}\textbf{83.7} & \color{blue}\textbf{86.5} & \color{blue}\textbf{$\text{88.08}$} \\
    \hline
    \rowcolor[gray]{0.9}\multicolumn{10}{c}{\textit{Adapter}~\cite{houlsby2019adapter}} \\ 
    $\text{Dyn-Adapter}_{0.7}$~\cite{zhang2024dynadapter}
    & 0.81 & 11.63 & 91.1 & 81.5 & 79.8 & 98.5 & 78.2 & 84.2 & 85.54 \\
    $\text{DMPO}_{0.7}$ (Ours)
    & 1.60 & 11.64 & \textbf{91.6} & \textbf{84.0} & \textbf{80.1} & \textbf{98.5} & \textbf{80.0} & \textbf{86.6} & \textbf{86.80} \\
    $\text{Dyn-Adapter}_{0.3}$~\cite{zhang2024dynadapter}
    & 0.81 & 5.18 & 72.6 & 77.3 & 66.3 & 98.4 & 60.8 & 54.8 & $\text{71.68}$ \\
    $\text{DMPO}_{0.3}$ (Ours)
    & 1.60 & 5.20 & \color{blue}\textbf{90.5} & \color{blue}\textbf{84.0} & \color{blue}\textbf{79.2} & \color{blue}\textbf{98.5} & \color{blue}\textbf{80.3} & \color{blue}\textbf{85.7} & \color{blue}\textbf{$\text{86.37}$} \\
    \hline
    \rowcolor[gray]{0.9}\multicolumn{10}{c}{\textit{Repadapter}~\cite{luo2023repadapter}} \\ 
    $\text{Dyn-Adapter}_{0.7}$~\cite{zhang2024dynadapter}
    & 1.67 & 11.61 & 91.7 & 83.1 & 80.7 & 98.0 & 83.3 & 85.2 & 87.02 \\
    $\text{DMPO}_{0.7}$ (Ours)
    & 0.89 & 11.62 & \textbf{92.3} & \textbf{86.6} & \textbf{81.3} & \textbf{99.2} & \textbf{84.1} & \textbf{87.2} & \textbf{88.44} \\
    $\text{Dyn-Adapter}_{0.3}$~\cite{zhang2024dynadapter}
    & 1.67 & 5.17 & 78.4 & 82.3 & 77.4 & 98.0 & 81.5 & 69.3 & $\text{81.14}$ \\
    $\text{DMPO}_{0.3}$ (Ours)
    & 0.89 & 5.19 & \color{blue}\textbf{91.7} & \color{blue}\textbf{86.5} & \color{blue}\textbf{80.7} & \color{blue}\textbf{99.2} & \color{blue}\textbf{84.1} & \color{blue}\textbf{86.3} & \color{blue}\textbf{$\text{88.08}$} \\
    \bottomrule
    \end{tabular}
    \caption{Results of different fine-tuning modules on CIFAR-100 and five FGVC datasets.}
    \label{tab:sup-fgvc}
\end{table*}

\section{Evaluation by Utilizing ImageNet-1K as Downstream Data}

\begin{figure}[t]
    \centering
    \includegraphics[width=0.4\textwidth]{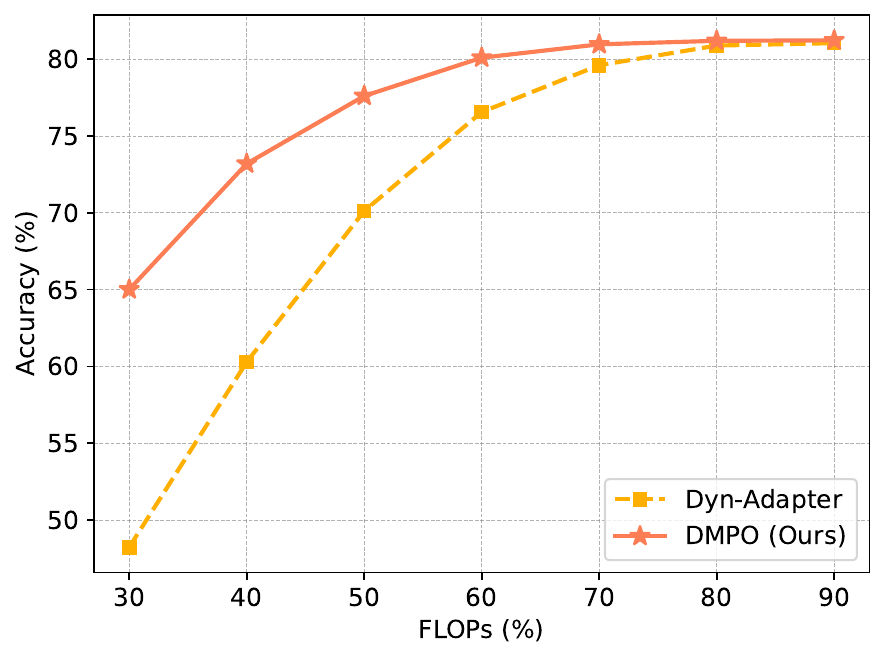}
    \caption{Results on ImageNet-1K at varying levels of inference FLOPs. Note that we use LoRA~\cite{hu2022lora} as the fine-tuning module and the results are obtained from training over 40 epochs. }
    \label{fig:imagenet}
\end{figure}

The results of Dyn-Adapter and our DMPO on ImageNet-1K after 40 epochs of training are shown in \cref{fig:imagenet}. DMPO consistently outperforms Dyn-Adapter across all inference FLOPs levels. Specifically, at 30\% FLOPs, DMPO achieves an accuracy of 63.76\%, significantly surpassing Dyn-Adapter's 48.23\% by a margin of 15.53\%. Overall, the accuracy curve of Dyn-Adapter declines sharply as FLOPs decrease, whereas DMPO demonstrates a much smoother decline. The accuracy gap between DMPO at 30\% and 70\% FLOPs is attributed to the complexity of the dataset and the limited number of training epochs.

\section{Ablation Studies}
To further validate the effectiveness of our DMPO, we conduct ablation studies on CIFAR-100 and five FGVC datasets. We use LoRA~\cite{hu2022lora} as the fine-tuning module.

\paragraph{Effect of Architecture Component}
\label{sec:rationale}
To further investigate the impact of different components of architecture, \ie, bypass module and high-order statistics-based predictor, we conduct ablation studies on both.
As shown in \cref{tab:architecture}, $\checkmark$ indicates that the corresponding component is used, while \ding{55} indicates that it is not used. From \cref{tab:architecture}, we can see that both architecture components contribute to improving performance.

\begin{table}[t]
  \centering
  \small
    \begin{tabular}{cc|cc}
    \toprule
    \multicolumn{2}{c|}{\textbf{Component}} & \multicolumn{2}{c}{\textbf{Average Acc (\%)}} \\
    \midrule
    \multicolumn{1}{c}{BYP} & \multicolumn{1}{c|}{HP} & 30\% FLOPs & 70\% FLOPs \\
    \midrule
    \ding{55}  & \ding{55}  & 84.41 & 86.96 \\
    \checkmark & \ding{55}  & 84.96 & 87.68 \\
    \ding{55}  & \checkmark & 87.76 & 88.21 \\
    \rowcolor[gray]{0.9} 
    \checkmark & \checkmark & \color{blue}\textbf{88.08} & \textbf{88.57} \\
    \bottomrule
    \end{tabular}%
\caption{Average results of DMPO with different architecture components. Note that \colorbox{gray!20}{gray} represents our DMPO.}
\label{tab:architecture}
\end{table}

\vspace{-1em}
\paragraph{Insertion Positions of High-order Statistics-based Predictor}
We further conduct experiments to evaluate the impact of inserting high-order statistics-based predictors at different stages. The results are presented in \cref{tab:mp_position}, where $\checkmark$ indicates the presence of a high-order statistics-based predictor at the corresponding position, while \ding{55} denotes its absence. As shown in \cref{tab:mp_position}, inserting a high-order statistics-based predictor at Stage 1 achieves better performance than at Stage 2, particularly at 30\% FLOPs. This is because, at 30\% FLOPs, most samples exit at Stage 1, where inserting a high-order statistics-based predictor allows to extract rich high-level discriminative features. However, at 70\% FLOPs, the features at Stage 1 are more low-level and representative compared to Stage 2. Without the high-order statistics-based predictor, the original predictor disrupts these low-level representative features more significantly, negatively impacting accuracy. Furthermore, using only one high-order statistics-based predictor in early stage results in a significant accuracy gap ($>1\%$) between 30\% FLOPs and 70\% FLOPs, but inserting high-order statistics-based predictors at both Stage 1 and Stage 2 effectively reduces this gap. Additionally, while inserting a high-order statistics-based predictor at Stage 3 additionally offers moderate performance improvements, the additional parameters and limited gains lead us to restrict insertions to Stages 1 and 2.

\begin{table}[t]
  \centering
  \small
    \begin{tabular}{cccc|c|cc}
    \toprule
    \multicolumn{4}{c|}{\textbf{Stage}} & \textbf{Param.} & \multicolumn{2}{c}{\textbf{Average Acc (\%)}} \\
    \cmidrule(lr){1-4} \cmidrule(lr){6-7}
    1 & 2 & 3 & 4 & \textbf{(M)} & 30\% FLOPs & 70\% FLOPs \\
    \midrule
    \ding{55}  & \ding{55}  & \ding{55} & \ding{55} & 1.00 & 84.96 & 87.68 \\
    \checkmark & \ding{55}  & \ding{55} & \ding{55} & 1.63 & 87.04 & 88.32 \\
    \ding{55}  & \checkmark & \ding{55} & \ding{55} & 1.34 & 85.42 & 87.66 \\
    \rowcolor[gray]{0.9} 
    \checkmark & \checkmark & \ding{55} & \ding{55} & 1.73 & 88.08 & \textbf{88.57} \\
    \checkmark & \checkmark & \checkmark & \ding{55} & 2.10 & \color{blue}\textbf{88.17} & 88.55 \\
    \bottomrule
    \end{tabular}%
\caption{Results of DMPO with different insertion positions of high-order statistics-based predictors. Note that \colorbox{gray!20}{gray} represents our DMPO.}
\label{tab:mp_position}
\end{table}

\begin{figure}
    \begin{subfigure}[ht]{0.235\textwidth}
        \centering
        \includegraphics[width=\textwidth]{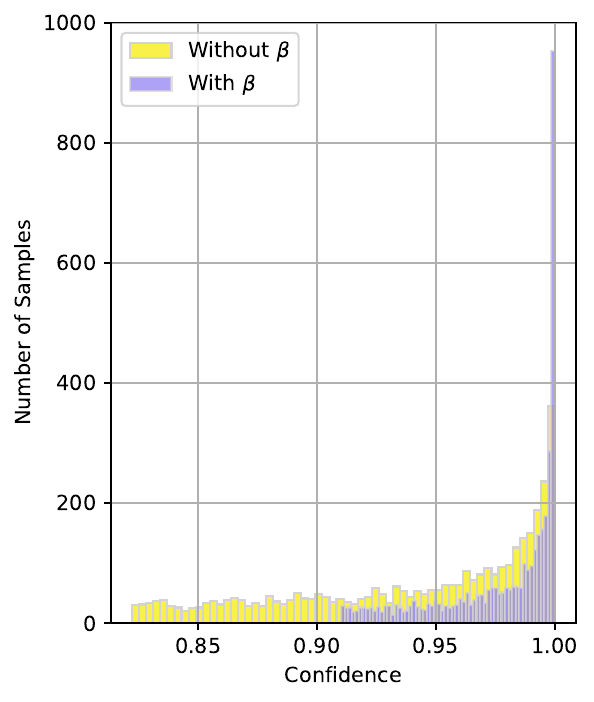}
        \caption{Visualization of confidence}
        \label{fig:confidence}
    \end{subfigure} 
    \hfill
    \begin{subfigure}[ht]{0.235\textwidth}
        \centering
        \includegraphics[width=\textwidth]{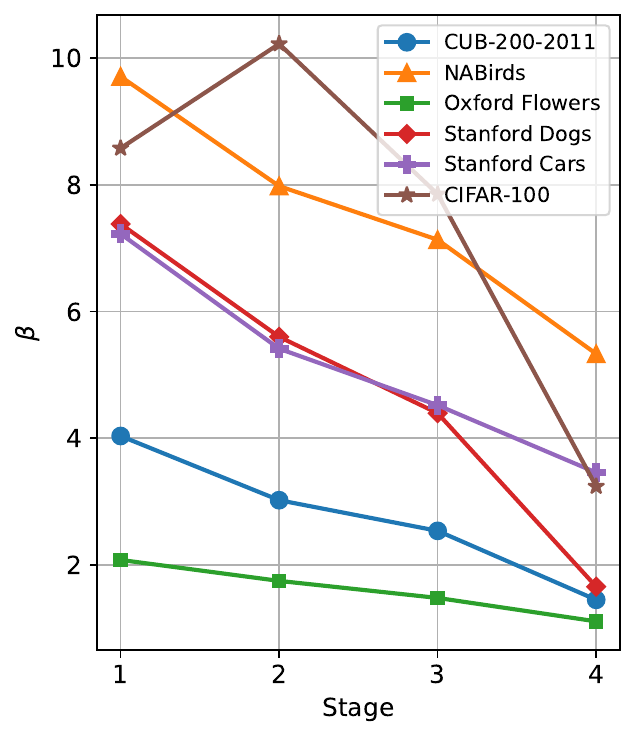}
        \caption{Final $\beta$ values for different stages}
        \label{fig:alpha_dif_stages}
    \end{subfigure}
    \caption{(a) Confidence distribution at Stage 1 with and without $\beta$ at 50\% inference FLOPs on CIFAR-100. The total number of samples is 3,500. (b) Final $\beta$ values across the four stages after training, with each stage's $\beta$ initially setting to 1.}
\end{figure}

\paragraph{Sensitivity of $\alpha$}
Regarding $\alpha$, results of different settings are summarized in \cref{tab:ablation_alpha}. It can be observed that $\alpha$ is not sensitive to different start and end values, and the default configuration is kept for different tasks. 

\begin{table}[h]
  \centering 
  \footnotesize
    \begin{tabular}{cccc}
    \toprule
    \multicolumn{1}{c}{\textbf{Start}} & \multicolumn{1}{c}{\textbf{End}} & \multicolumn{1}{c}{\textbf{Acc\textsubscript{0.3}}} & \multicolumn{1}{c}{\textbf{Acc\textsubscript{0.7}}} \\
    \midrule
    \rowcolor[gray]{0.9} $[0.01, 0.01, 1.0, 2.0]$ & $[1.5, 1.0, 0.1, 0.1]$ & 88.08 & 88.57 \\
    $[0.01, 0.01, 2.0, 5.0]$ & $[2.0, 1.0, 0.1, 0.1]$ & 88.04 & 88.45 \\
    $[0.01, 0.01, 0.5, 1.5]$ & $[2.0, 1.0, 0.1, 0.1]$ & 88.08 & 88.45 \\
    $[0.01, 0.01, 2.0, 5.0]$ & $[5.0, 2.0, 0.1, 0.1]$ & 88.05 & 88.40 \\
    \bottomrule
    \end{tabular}%
    \setlength{\abovecaptionskip}{3pt} 
    \setlength{\belowcaptionskip}{3pt}
    \caption{Results of various $\alpha$ settings. \colorbox{gray!20}{Gray} is default setting.}
\label{tab:ablation_alpha}
\end{table}

\paragraph{Effect of $\beta$}
To validate the effectiveness of $\beta$ on performance, we conduct an ablation study. As shown in \cref{tab:beta_ablation}, \ding{55} indicates that $\beta$ is not used, while \checkmark denotes its use. The results demonstrate that $\beta$ can further enhance overall performance by improving the confidence of early stages in classification. Besides, \cref{fig:confidence} proves that learned $\beta$ effectively enhances the classification confidence of early stages. \cref{fig:alpha_dif_stages} illustrates that the values of learned $\beta$ generally decrease with increasing stage depth across most datasets. This trend reveals that, compared to deeper stages, early stages exhibit lower classification confidence due to limited discriminative ability. Consequently, early stages tend to learn larger $\beta$ values to enhance confidence.

\begin{table}[ht]
  \centering
  \small
  \setlength{\tabcolsep}{1.1em}
    \begin{tabular}{l|cc}
    \toprule
    \multirow{2}{*}{\textbf{$\beta$}} & \multicolumn{2}{c}{\textbf{Average Acc} (\%)} \\
\cmidrule{2-3}          & 30\% FLOPs  & 70\% FLOPs \\
    \midrule
    \ding{55} & 87.08 & 87.76  \\
    \rowcolor[gray]{0.9} \checkmark & \color{blue}\textbf{88.08} & \textbf{88.57}  \\
    \midrule    
    \end{tabular}%
      \caption{Average results with and without $\beta$. Note that \colorbox{gray!20}{gray} represents our DMPO.}
  \label{tab:beta_ablation}%
\end{table}%

\begin{figure}[ht]
  \centering
  \includegraphics[width=0.4\textwidth]{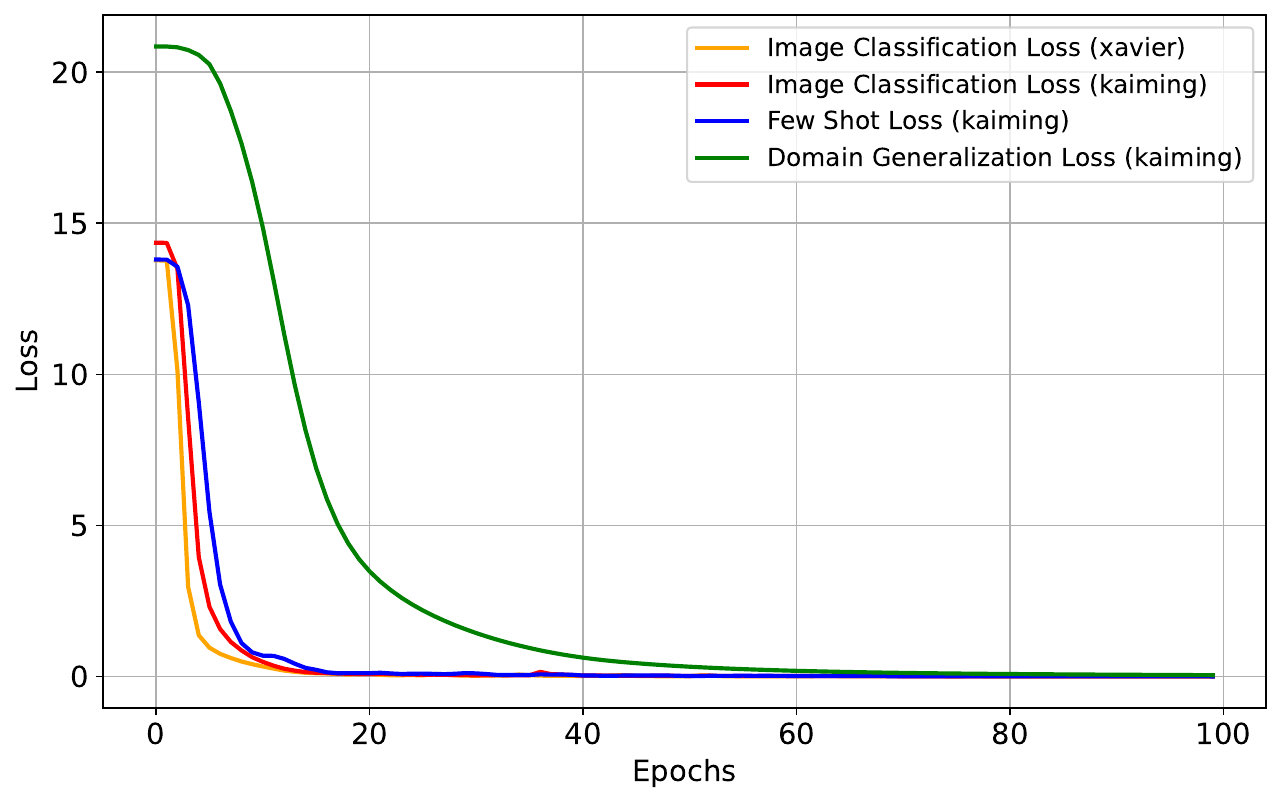}
   \caption{Training losses across tasks and initialization methods. }
   \label{fig:loss}
\end{figure}
\section{Convergence of Decoupled Optimization}
As illustrated by the loss curves in \cref{fig:loss}, our decoupled optimization algorithm demonstrates stable and rapid convergence across various tasks and parameter initialization methods.

\section{FLOPs of Different Module}
For different model parts per stage, FLOPs is recorded as 0.0024G (1 Bypass), 0.048G (1 HP classifier),  being much lower than 4.2G (3 ViT blocks) and leading negligible cost.

\section{Training Overhead}
We evaluate the training time and memory overhead with a batch size of 32 on a single NVIDIA 3090 GPU. The comparison of training samples per seconds is: 221 images/s (LoRA) \emph{v.s.} 218 images (Dyn-Adapter) \emph{v.s.} 194 images (DMPO). The memory consumption is: 5.55GB (Dyn-Adapter) \emph{v.s.} 5.67GB (DMPO). Overall, these results indicate that DMPO introduces minimal additional training overhead.

\section{Visualization of Images at Different Stages}
\begin{figure*}[hp]
    \centering
    \includegraphics[width=0.85\textwidth]{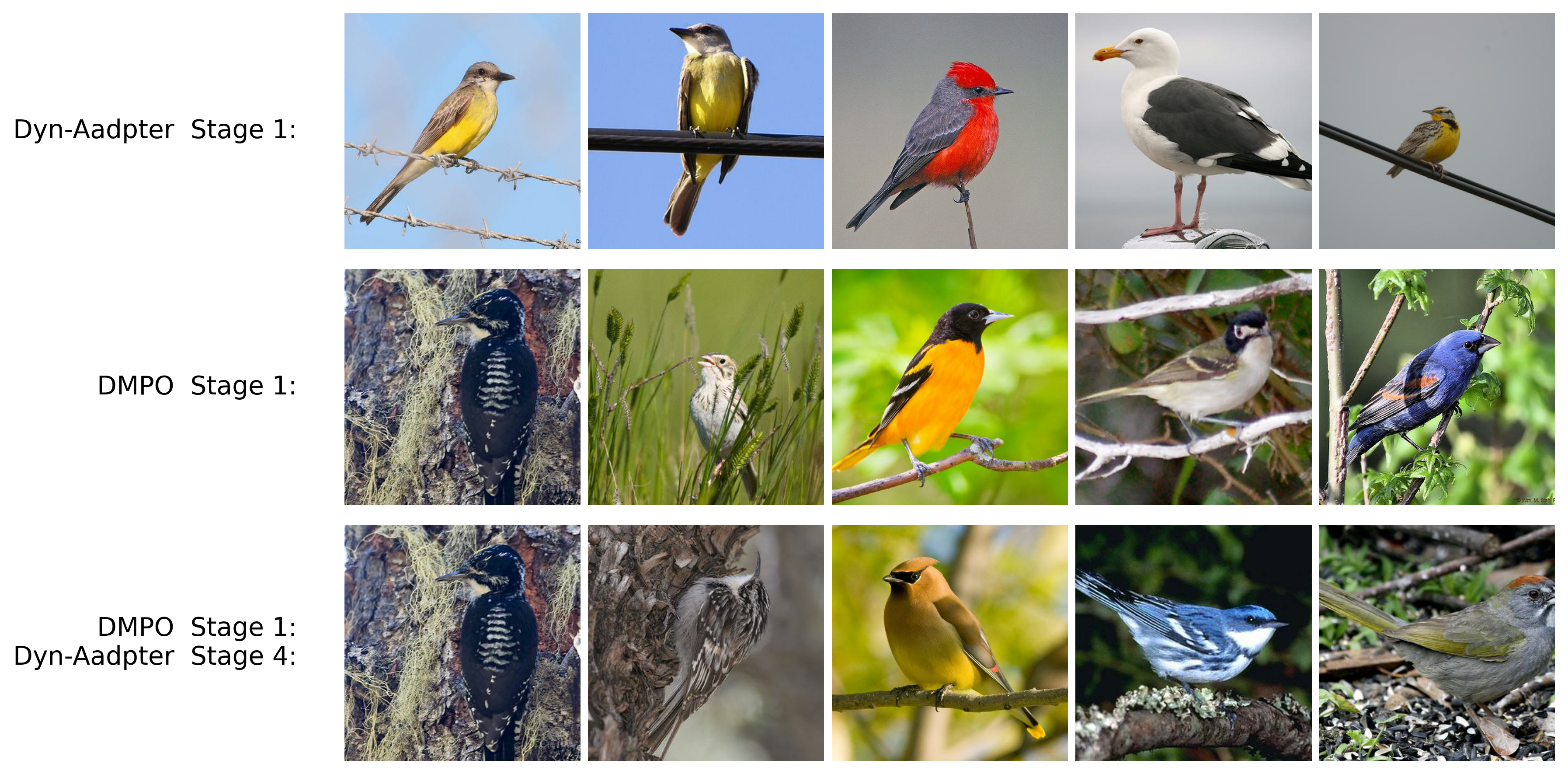}
    \caption{Visualization of the classified images at different stages by our DMPO and Dyn-Adapter. The top row shows images correctly classified by Dyn-Adapter at Stage 1, the middle row shows images correctly classified by DMPO at Stage 1, and the bottom row presents images correctly classified by DMPO at Stage 1 but only correctly classified by Dyn-Adapter at Stage 4.}
    \label{fig:images}
    \vspace{128in}
\end{figure*}

To intuitively demonstrate the feasibility of our DMPO, we visualize the classified images at different stages. As shown in \cref{fig:images}, we train Dyn-Adapter~\cite{zhang2024dynadapter} and our DMPO on the CUB-200-2011 dataset and select five correctly classified images from Stage 1 and Stage 4 at 70\% FLOPs. We use LoRA~\cite{hu2022lora} as the fine-tuning module. From \cref{fig:images}, it can be observed that our DMPO outperforms Dyn-Adapter in correctly classifying more complex images at early stages, indirectly indicating that DMPO's early stages exhibit stronger discriminative ability compared to Dyn-Adapter.

\end{document}